\documentclass{article} 
\usepackage{iclr2026_conference,times}

\usepackage{amsmath,amsfonts,bm}

\def\eqref#1{equation~\ref{#1}}

\def\1{\bm{1}}

\DeclareMathAlphabet{\mathsfit}{\encodingdefault}{\sfdefault}{m}{sl}
\SetMathAlphabet{\mathsfit}{bold}{\encodingdefault}{\sfdefault}{bx}{n}

\usepackage{subcaption} 
\usepackage{url}
\usepackage{graphicx} 
\usepackage[hidelinks]{hyperref}
\usepackage{fontawesome5}
\usepackage{caption}
\usepackage{booktabs,longtable,threeparttable,tabularx,pifont,xcolor}
\usepackage{listings}
\newcolumntype{Y}{>{\raggedright\arraybackslash}X}
\newcommand{\cmark}{\textcolor{green!60!black}{\ding{51}}}
\newcommand{\xmark}{\textcolor{red!70!black}{\ding{55}}}

\usepackage[table]{xcolor} 
\usepackage{arydshln}
\newcommand{\dmidrule}{%
  \addlinespace[0.3ex]%
  \arrayrulecolor{black!40}\hdashline[0.4pt/2pt]\arrayrulecolor{black}%
  \addlinespace[0.3ex]%
}

\usepackage{multirow}
\usepackage{array,tabularx}
\newcolumntype{C}{>{\centering\arraybackslash}X}

\usepackage{ragged2e}
\newcolumntype{P}[1]{>{\RaggedRight\arraybackslash}p{#1}}
\newcolumntype{Q}[1]{>{\Centering\arraybackslash}p{#1}}

\definecolor{msOrange}{HTML}{F65314}
\definecolor{msGreen}{HTML}{7CBB00}
\definecolor{msBlue}{HTML}{00A1F1}
\definecolor{msYellow}{HTML}{FFBB00}

\usepackage[T1]{fontenc}
\usepackage{sourcesanspro} 

\iclrfinalcopy

\newcommand{\tok}[2]{\colorbox{#1}{\strut #2}}

\lstdefinestyle{prompt}{
  basicstyle=\rmfamily\itshape\small, 
  columns=fullflexible,
  keepspaces=true,
  escapeinside={(*@}{@*)},
  showstringspaces=false,
  upquote=true
}

\usepackage{pgf} 
\usepackage{xfp}
\usepackage{tikz}
\usepackage{fancyhdr}
\usepackage{amssymb}
\usepackage{tikzpagenodes}
\AddToHook{shipout/firstpage}{%
  \begin{tikzpicture}[remember picture,overlay]
    \node[anchor=south west, inner sep=0pt]
      at ([xshift=1.5in,yshift=0.6in]current page.south west)
      {\scriptsize \textsuperscript{\dag}\texttt{At present, \textsc{MachineLearningLM} targets \emph{tabular ML}; broader ML is future work.}};
  \end{tikzpicture}}
  \AddToHook{shipout/firstpage}{%
  \begin{tikzpicture}[remember picture,overlay]
    \node[anchor=south west, inner sep=0pt]
      at ([xshift=1.5in,yshift=0.46in]current page.south west)
      {\scriptsize \textsuperscript{*}\texttt{Corresponding author; details in Appendix~\ref{sec:author}.}};
  \end{tikzpicture}}
  
\title{MachineLearningLM\footnotemark[2] : Scaling Many-Shot \\\scalebox{0.94}{In-Context Learning via Continued Pretraining}}

\author{Haoyu Dong\textsuperscript{*}, Pengkun Zhang, Mingzhe Lu, Yanzhen Shen, Guolin Ke\\
UCAS, Microsoft, SCUT, Stanford 
}

\begin{document}

\vspace*{-0.8cm}
\maketitle
\vspace*{-0.4cm}

\begin{abstract}
Large language models (LLMs) possess broad world knowledge and strong general-purpose reasoning ability, yet they struggle to learn from \emph{many} in-context examples on standard machine-learning (ML) tasks—i.e., to leverage many-shot demonstrations purely via in-context learning (ICL) without gradient descent. We introduce \textsc{MachineLearningLM}, \emph{a portable continued-pretraining framework} that equips a general-purpose LLM with \emph{robust many-shot ICL capability} while preserving its general knowledge and reasoning for broader chat workflows.

Our pretraining procedure synthesizes ML tasks from \emph{millions of structural causal models (SCMs)}, spanning shot counts up to $1{,}024$. We begin with a random-forest teacher, distilling tree-based decision strategies into the LLM to strengthen robustness in numerical modeling. All tasks are serialized with a \emph{token-efficient prompt}, enabling \emph{$3$–$6\times$} more examples per context window and delivering up to \emph{$50\times$} amortized throughput via batch inference.

Despite a modest setup (\emph{Qwen-2.5-7B-Instruct} with LoRA rank~8), \textsc{MachineLearningLM} \emph{outperforms strong LLM baselines} (e.g., GPT-5-mini) by an average of $\sim$15\% on out-of-distribution tabular classification across finance, physics, biology, and healthcare domains. It exhibits a \emph{striking many-shot scaling law}: accuracy increases monotonically as in-context demonstrations grow from $8$ to $1{,}024$. Without any task-specific training, it attains \emph{random-forest–level} accuracy across hundreds of shots. General chat capabilities—including knowledge and reasoning—are preserved: it achieves 75.4\% on MMLU.
\end{abstract}

\vspace{-1.4em}
\begin{center}
\small

\href{https://huggingface.co/MachineLearningLM}{%
  \raisebox{-0.2ex}{\includegraphics[height=1.1em]{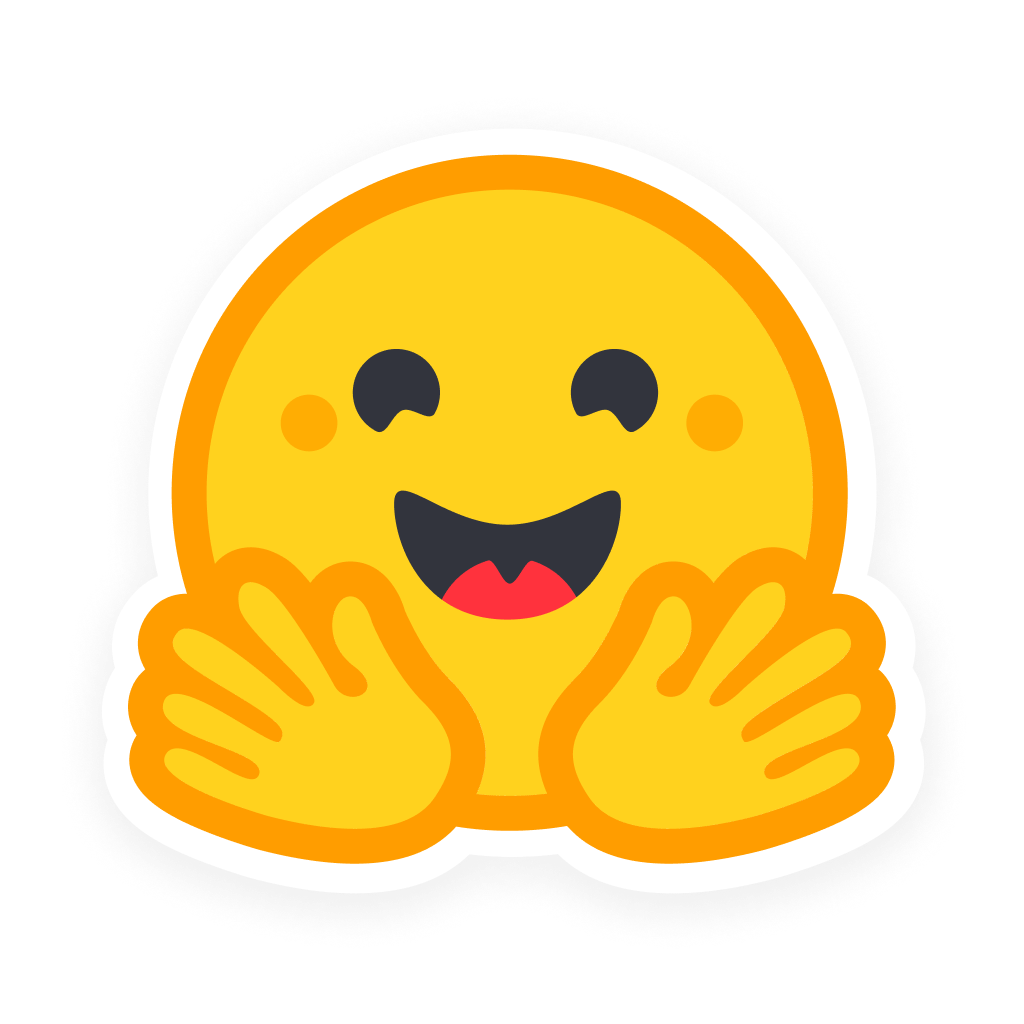}}\ \textbf{Model \& Data}: \url{https://huggingface.co/MachineLearningLM}}\\
\href{https://github.com/HaoAreYuDong/MachineLearningLM}{\faGithub\ \textbf{Code}: \url{https://github.com/HaoAreYuDong/MachineLearningLM}}

\end{center}

\vspace{-0.5em}
\begin{center}
  \includegraphics[width=\linewidth]{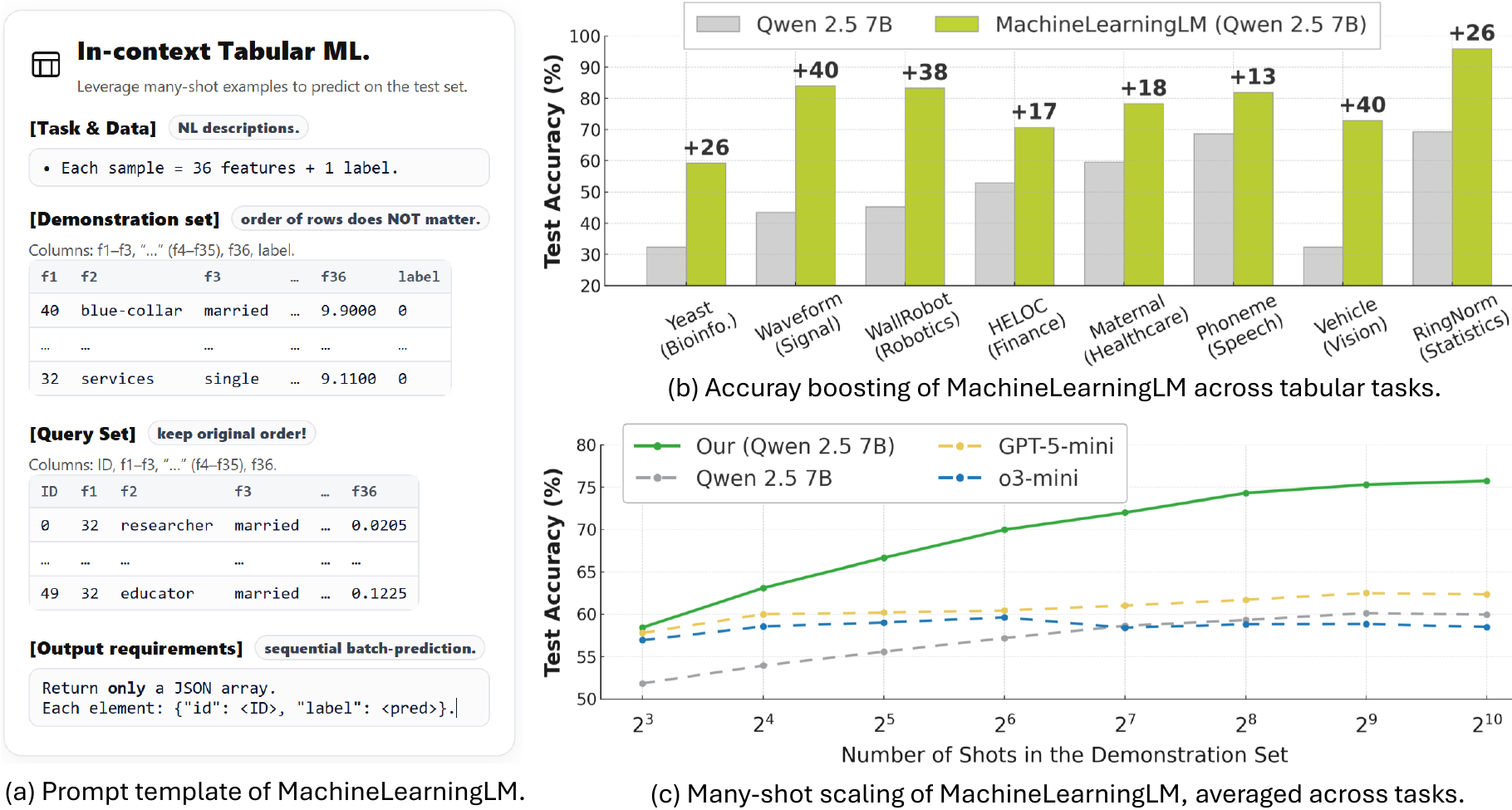}
  \captionof{figure}{\textsc{MachineLearningLM} on in-context ML tasks: (a) prompt template; (b) 512-shot accuracy across domains vs.\ Qwen-2.5-7B-Instruct; (c) many-shot scaling ($2^3$–$2^{10}$ shots) vs.\ LLMs.}\label{fig:cover}
\end{center}
\vspace{-0.6em}

\section{Introduction}\label{sec:intro}

LLMs possess broad world knowledge, general perceptual as well as reasoning abilities, making them promising few-shot learners~\citep{brown2020language}. However, they often fail to learn new tasks despite being given many-shot demonstrations on standard ML benchmarks~\citep{agarwal2024manyshot,gardner2024largescale}, or to fully exploit rich experiences stored in agent memory in interactive, open-ended workflows~\citep{wang2023voyager,shinn2023reflexion,wang2024agent}. In addition, accuracy gains typically plateau —often after just a handful of demonstrations— and are sensitive to the label biases and choice/order of examples ~\citep{chen2023howmany,liu2024lostinthemiddle,zhao2021calibrate,fei2023mitigating,liu2022makes}. In practice, LLMs are largely guided by surface-level signals, such as distributional, formatting cues~\citep{min-etal-2022-rethinking} and nearest-neighbor imitation~\citep{agarwal2024manyshot}, and rarely uncover new causal mechanisms or statistical dependencies that are required to yield accurate predictions.

Orthogonally, pioneering tabular models~\citep{hollmann2025accurate,qu2025tabicl} have demonstrated ML tasks can be solved purely by ICL—\emph{without gradient descent}. However, these tabular-specific model architectures cannot leverage the broad prior world knowledge and general-purpose multimodal perception that LLMs acquire during pretraining; consequently, they depend heavily on well-designed featurization~\citep{shi2021benchmarking,mraz2025towards} and abundant labeled data for training. At this intersection, we ask:\\ 
\vspace{-1.2em}
\begin{center}
\normalsize \bfseries\itshape
Can we teach an LLM to ``do ML in context'' while preserving general abilities?
\end{center}

To this end, we propose a \emph{pretraining-plus-prompting} framework, \textsc{MachineLearningLM}, that equips LLMs with in-context ML capabilities to fully exploit many-shot in-context examples. \textsc{MachineLearningLM} performs LoRA-based~\citep{hu2022lora} continued pretraining with a standard next-token objective on millions of \emph{synthetic} tabular prediction tasks drawn from SCM-based priors~\citep{peters2017elements,pearl2009causality}: a task generator samples arbitrarily large numbers of binary/multiclass tasks under SCM-based priors spanning diverse feature types, marginal distributions, and label mechanisms following~\citep{qu2025tabicl}. This approach ensures strict non-overlap between our pretraining data and evaluation datasets. After pretraining, the model can directly leverage in-context examples of a new task to generate predictions for unseen instances—\emph{without} parameter update.

As summarized in Table~\ref{tab:comp}, unlike previous instruction-tuning methods \citep{wang-etal-2022-super,chung2022scaling} that rely on a limited set of real tasks (typically \(\sim 10^3\)), we train on \(\mathcal{O}(10^6)\) synthetic tasks with \emph{diverse causal mechanisms and varied shot counts}. Our approach also differs from specialized tabular learners as we preserve the versatility of LLMs, which enables them to continue to leverage contextual task descriptions, draw on external knowledge, and interact directly with multimodal, heterogeneous inputs. Through large-scale pretraining, our method is able to equip LLMs with striking many-shot scaling and robust numerical modeling. We hope \textsc{MachineLearningLM} is a promising paradigm to inspire new research that leverages LLMs' general perception and reasoning capabilities, and also possibly extending to more modalities beyond text, as detailed in Section~\ref{future}.

\begin{table}[!hb]
\begingroup
\footnotesize 
\centering
\setlength{\tabcolsep}{6pt}
\renewcommand{\arraystretch}{1.2}

\newcolumntype{Y}{>{\raggedright\arraybackslash}X}
\newcolumntype{C}{>{\centering\arraybackslash}X}

\definecolor{YesColor}{HTML}{8DC63F} 
\definecolor{NoColor}{HTML}{B82B2B}  

\providecommand{\cmark}{\ding{51}}
\providecommand{\xmark}{\ding{55}}
\renewcommand{\cmark}{\textcolor{YesColor}{\ding{51}}}
\renewcommand{\xmark}{\textcolor{NoColor}{\ding{55}}}

\providecommand{\dmidrule}{\midrule}

\begin{threeparttable}
\caption{Comparison across ML paradigms.}
\begin{tabularx}{\linewidth}{@{}YCCCCC@{}} 
\toprule
\textbf{Method} &
\textbf{In-context learning} &
\textbf{Robust numerical modeling} &
\textbf{General knowledge priors} &
\textbf{Native multimodal input} \\
\midrule
RF, Boosted trees        & \xmark & \cmark & \xmark & \xmark \\
\dmidrule
TabPFN, TabICL            & \cmark & \cmark & \xmark & \xmark \\
\dmidrule
TabLLM                    & \xmark & \xmark & \cmark & \cmark \\
\dmidrule
GPT-5, Qwen           & \cmark & \xmark & \cmark & \cmark \\
\dmidrule
\textsc{MachineLearningLM}& \cmark & \cmark & \cmark & \cmark \\
\bottomrule
\end{tabularx}
\label{tab:comp}
\begin{tablenotes}[para,flushleft]
\footnotesize
\textbf{Key}: \cmark\;Yes, \xmark\;No. \;
\textbf{Note}: ``Native multimodal'' primarily means \emph{textual + numerical + tabular} and can be naturally extended to modalities like images by building on multimodal LLM backbones. 
\end{tablenotes}
\end{threeparttable}
\endgroup
\end{table}


\paragraph{Warm-up training with a Random Forest teacher.}
Directly training on synthetic tasks can lead to model collapse or underperformance, particularly when tasks are too complex to be learn from only a few (e.g., 64) examples. This makes strong ML methods no better than random guessing or always predicting the majority class. We stabilize the onset by \emph{mimicking a random-forest (RF) teacher} on each task—first matching example predictions—before transitioning to self-reliant in-context prediction. This leverages knowledge distillation for improved optimization~\citep{hinton2015distillation}. We choose a random forest teacher for its balance of robustness and interpretability. Its decision process can be transparently decomposed into rule paths and feature attributions and directly serialized into interpretable ``reasoning steps''—–then rule chains and faithful local explanations~\citep{deng2014intrees,friedman2008rulefit,lundberg2017shap}, which naturally align with the chain-of-thought (CoT) reasoning. In this work, we distill the predicted labels only, but future directions could leverage its ``reasoning steps'' as rationales for reasoning-augmented training~\citep{deepseek-r1} and enhancing the interpretability of model predictions.

\paragraph{Token-efficient prompting for in-context ML.}
LLMs are fundamentally bottlenecked by their context length. Our design \emph{enables 3-6$\times$} more examples within a context window and yields \emph{$50\times$} amortization via batch inference. We achieve this with three composable design choices.

(i) A \emph{tabular encoding} to organize many-shot examples rather than as \emph{scattered and lengthy} NL descriptions~\citep{hegselmann2023tabllm,gardner2024largescale}.   
Recent works~\citep{dong2024spreadsheetllm,sui2023tablemeetsllm} have shown that regular tabular structures can be seamlessly understood by LLMs without requiring special row/column attention~\citep{hollmanntabpfn}. In addition, tabular encoding~\footnote{It is extensible and can support images, hyperlinks, and embedded objects, as in HTML tables.} naturally combines \emph{heterogeneous} with no mandatory categorization for text.

(ii) A \emph{compact integer-based encoding} in which we normalize all numbers for each numerical feature to integers in $[0,999]$, eliminating the tokenization fragmentation caused by decimal points (``.'') and leading signs (``+''/``–''). This not only reduces the token count—e.g., \texttt{cl100k\_base} treats any integer in $[0,999]$ as a single token, but also avoids a common pitfall where LLMs compare decimals as strings (e.g., ``1.11'' \texttt{(1|.|11)} vs.\ ``1.9'' \texttt{(1|.|9)})~\citep{hfnumtokblog2024}.  

(iii) A \emph{sequence-level batch-prediction} mechanism that packs dozens of test examples into each sequence and predicts all of them in a single forward pass, stabilizing gradients for continued pretraining and amortizing instruction/context overhead~\citep{lin2023batchprompt,cheng2023batch}.   

\paragraph{Order-robust, confidence-aware self-consistency.}
LLM predictions can be sensitive to long-context position effects~\citep{liu2024lostinthemiddle}. At inference, we marginalize over permutations of demonstration and feature order, combining outputs via \emph{confidence-weighted self-consistency}—i.e., a calibrated, probability-weighted vote over samples~\citep{zhao2021calibrate,wang2022selfconsistency}. 

\textsc{MachineLearningLM} continues pretraining on millions of synthetic tasks with diverse causal mechanisms and varied shot counts; our \textbf{key experimental highlights} are:

\begin{itemize}

  \item \textbf{Many-shot scaling.}
  \textsc{MachineLearningLM} exhibits striking many-shot scaling across diverse tasks from the TALENT benchmark~\citep{liu2024talenttabularanalyticslearning}, surpassing both leading open-source (e.g. Qwen-2.5-7B) and closed-source (e.g. GPT-5-mini) LLMs by an average of $\sim$15\% under high-shot settings. Moreover, it exhibits clear out-of-distribution generalization in context length, from a 32k-token pretraining budget to a 131k-token inference budget.

\item \textbf{Competitive many-shot performance vs.\ state-of-the-art tabular methods.}
Without any task-specific training, \textsc{MachineLearningLM} reaches \emph{random-forest–level} accuracy across shot counts from 8 to 512 (within \textbf{2\%} \emph{relative} on average) and clearly surpasses instance-based kNN, demonstrating robust numerical modeling.

\item \textbf{Heterogeneous (multimodal)-input generalization.}
By combining LLM versatility with classical ML’s robust numerical fitting, \textsc{MachineLearningLM} natively consumes natural-language (NL) features alongside numerical values—\emph{without relying on text bucketing or embeddings}—and achieves competitive results on the Talent benchmarks.

  \item \textbf{General abilities preserved (chat workflows).}
  \textsc{MachineLearningLM} retains general capabilities in chat-style workflows; for example, on MMLU it attains 73.2\% micro accuracy in 0-shot and 75.4\% in 50-shot, comparable to strong general-purpose LLMs.

\end{itemize}

\section{Method}

\subsection{Pretraining Corpus Synthesis}
\label{synthesis_data}

To generate pretraining data, we build a task generator following TabICL\footnote{We directly use the source code of TabICL in \url{https://github.com/soda-inria/tabicl}.}~\citep{qu2025tabicl} that samples diverse binary/multiclass problems under SCM-based priors determined by choices of graph structure, mechanisms, feature types, and class formation.

\paragraph{SCM graph and mechanisms.}
We first sample a DAG $\mathcal{G}$ following the structure of a fully connected MLP, where each neuron corresponds to a variable, then assign structural equations to each node $v$ in $\mathcal{G}$:
\[
x_v \;\leftarrow\; f_v\!\big(x_{\mathrm{parents}(v)}\big) \;+\; \varepsilon_v,\qquad \varepsilon_v \ \text{i.i.d. noise}.
\]
where $f_v$ is drawn from a rich pool of functions, following \citep{qu2025tabicl}, including \emph{tanh, leaky-ReLU, ReLU, ReLU6, SELU, SiLU, Softplus, Hardtanh, sign, sine, RBF, exp, gaussian-process-sampled random activations}, 

\paragraph{Tree-based SCMs.}
To inject tree inductive biases beyond neural layers, 30\% $f_v$s are replaced by multi-output gradient-boosted regressors ($f_\ell$) fitted on fake targets drawn from Gaussian noise and parent inputs. For each layer:

\[
\begin{aligned}
n_{\text{estimators}} &\;\sim\; \min\!\bigl\{4,\; 1 + \mathrm{Exponential}(\lambda=0.5)\bigr\}, \\[6pt]
\text{max\_depth} &\;\sim\; \min\!\bigl\{4,\; 2 + \mathrm{Exponential}(\lambda=0.5)\bigr\}, \\[6pt]
\text{We also fit } f_\ell &: \; x_{\mathrm{parents}(\ell)} \;\mapsto\; y 
   \quad \text{with } y \sim \mathcal{N}(0,I), \\[6pt]
\text{and set } x_\ell &\;\leftarrow\; f_\ell\!\bigl(x_{\mathrm{parents}(\ell)}\bigr).
\end{aligned}
\]

\paragraph{Feature and label typing.}
We sample a fraction of categorical features, discretize dense columns into bins, and (with some probability) shuffle category IDs to avoid spurious order. Continuous features remain numeric. As for a scalar $y^\star$. We create $K$-way labels by sampling $K\!\in\!\{2,\ldots,10\}$, drawing ordered bounds $\tau_1{<}\cdots{<}\tau_{K-1}$ from the empirical distribution of $y^\star$, and mapping
\[
y \;=\; \arg\max_{k\in\{1,\ldots,K\}} \mathbb{1}\!\big[\tau_{k-1} \le y^\star < \tau_k\big],
\]

Followed by a random permutation of class IDs to destroy ordinal hints. Class imbalance arises naturally from the sampled bounds.

\begin{figure}[!t]
  \centering
  \includegraphics[width=0.78\linewidth]{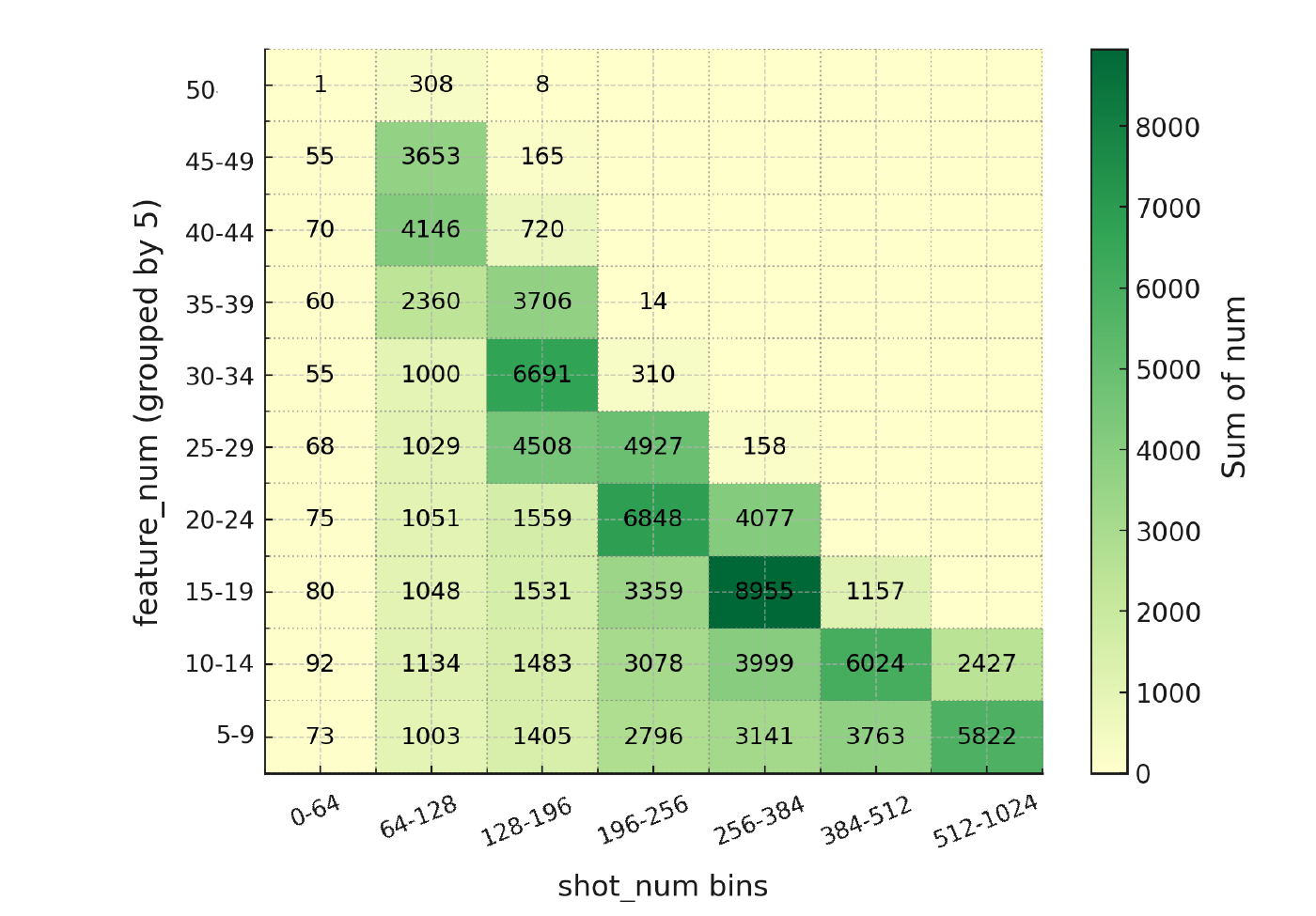}
  \vspace{-0.6em}
  \caption{Heatmap of task density across feature and shot counts. We sample 100k tasks from the synthetic generator and aggregate by \texttt{feature\_num} and \texttt{shot\_num} bins. Color encodes the sum of \texttt{num} per bin. Training token budget capped at 32k tokens.}
  \label{fig:heatmap}
\end{figure}

\paragraph{Task sampling policy and corpus scale.}
We repeatedly instantiate fresh SCMs, yielding a corpus of 
\(\sim 3\times 10^6\) distinct SCMs. For each SCM, we propagate noise through the structural equations to generate i.i.d.\ samples \((X, y)\). For each task, we sample the number of demonstrations \(M\) (capped at \(1{,}024\)), fix the number of queries at \(N=50\), and draw the number of features \(d \in \{5,\dots,50\}\) and classes \(K \in \{2,\dots,10\}\).
Each task is serialized into a prompt following Section ~\ref{sec:prompting} and comprises the following components:
a compact tabular many-shot block (features with labels), a tabular test block (IDs with features),
and an instruction specifying a JSON array output \(\big[\{\texttt{"id"}:\cdot,\texttt{"label"}:\cdot\},\ldots\big]\), as illustrated in Figure~\ref{fig:cover}a.
Given a token budget of \(32\text{k}\) Qwen~2.5-Instruct in our framework, we truncate the number of many-shot samples per task to fit the length limitation for continued pre-training.

\subsection{Objective of Continued Pre-training}
\label{sec:objective}
As illustrated in Figure~\ref{fig:cover}a, we cast each in-context tabular prediction task as a \emph{conditional sequence modeling} problem. For a given task, the prompt \(x\) concatenates: (i) an instruction \emph{header} \(H\) describing the task and specifying a JSON output format, (ii) a many-shot training block with \emph{a demonstration block} \(S=\{s_i\}_{i=1}^{M}\) of \(M\)-shot labeled demonstrations (demonstration order immaterial), and (iii) \emph{a query block} \(Q=\{x^{(i)}\}_{i=1}^{N}\) of \(N\) unlabeled queries (each with an \texttt{id}). (ii) and (iii) follow the token-efficient prompting in Sec.~\ref{sec:prompting}. 
\[
\underbrace{H}_{\text{instruction + schema}}
\;\Vert\;
\underbrace{s_{1} \,\Vert\, \cdots \,\Vert\, s_{i} \,\Vert\, \cdots \,\Vert\, s_{M-1}}_{\text{in-context demonstrations}}
\;\Vert\;
\underbrace{x_{1} \,\Vert\, \cdots \,\Vert\, x_{N-1}}_{\text{batched queries}}
\]
The model is required to generate a single JSON array.
\[
y \;=\; \big[\{\texttt{"id"}:0,\;\texttt{"label"}:\hat{\ell}_1\},\ldots,\{\texttt{"id"}:N -1,\;\texttt{"label"}:\hat{\ell}_{N-1}\}\big].
\]

\paragraph{Loss.}
We introduce no auxiliary heads or bespoke objectives, and use the standard left-to-right log-likelihood language modeling objective \emph{over the JSON target} for continued pre-training. Formally, let \(\mathrm{tok}(\cdot)\) denote the tokenizer and
\[
\mathbf{x}=\mathrm{tok}(x),\qquad 
\mathbf{y}=\mathrm{tok}(y)=(y_1,\ldots,y_T).
\]
The training loss for parameters \(\theta\) is the negative log-likelihood
\begin{equation}
\mathcal{L}(\theta)
\,=\,
-\sum_{t=1}^{T}\log p_{\theta}\!\left(y_t \mid \mathbf{x}, y_{<t}\right),
\label{eq:lml}
\end{equation}
applied to the \emph{entire} JSON string (brackets, keys, colons, commas, IDs, and label tokens). This encourages (i) learning the mapping from features to labels, (ii) preserving test-row order via aligned \texttt{"id"} fields, and (iii) strict adherence to the required JSON format.

\subsection{Random-Forest Mimic Warm Start}
Directly training on synthetic tasks can cause training collapse as low-signal tasks, limited few-shot examples, severe class imbalance, and randomly sampled label regimes may yield poor local optima, bend the loss curve, and occasionally destabilize training \citep{ochal2023few}. We mitigate this risk by first \emph{mimicking a Random Forest (RF) teacher} on each task, where we match per-example predictions in a short warm-up—before transitioning to standalone in-context prediction. This provides more informative targets and smoother gradients in the early phase, similar to knowledge distillation~\citep{hinton2015distillation}. We choose RF as the teacher because its decision process decomposes transparently into rule paths and feature attributions. These can be serialized as explicit ``reasoning steps''~\citep{deng2014intrees,friedman2008rulefit,lundberg2017shap}, which aligns naturally with the stepwise reasoning patterns large language models can emulate.

For each pretraining task, we (i) utilize a \emph{better-than-random guard} at the task level to skip low-signal or degenerate tasks, and (ii) apply an \emph{example-level consensus filter} that retains only test examples where the RF prediction matches the ground truth; we filter examples at this stage but \emph{do not} alter the original train/test labels. After the warm-up ends, we continue to apply (i) but \emph{disable} (ii), enabling the model to transition from relying on random-forest–induced priors to developing its own in-context ML behavior. This design prunes low-signal tasks before they corrupt optimization and, by supplying high-precision targets, reduces gradient noise, prevents collapse-to-majority under imbalance, and yields a smoother warm-up that mimics robust numerical modeling—ultimately easing the shift from teacher imitation to independent in-context ML.

\paragraph{(i) Task-level guard against chance and collapse.}
To rule out degenerate tasks where the RF performs no better than chance, we compare its accuracy to a conservative random baseline
\[
p_0 \;=\; \max\!\Big(\sum_{k=1}^{K} p_k^2,\; \max_k p_k\Big),
\]
where $K$ is the number of classes, and $p_k$ denotes the class prior from $y_{\text{true}}$. This baseline corresponds to the larger of ``sampling by prior'' or ``always predicting the majority class.''
Let $N_{correct}=\#\{i:y_i=\hat{y}_i\}$ be the number of correct predictions, we then perform a one-sided binomial tail test
$p\text{-value}=\Pr[X\!\ge\!N_{correct}],\;X\!\sim\!\mathrm{Bin}(N,p_0)$ and require $p\text{-value}<\alpha$.\footnote{For $N=50$ the test has relatively low power; we use a slightly relaxed $\alpha$ (default $0.2$) to avoid discarding borderline-but-promising tasks.}

To avoid trivial solutions and collapsed behavior under class imbalance, we enforce the following criteria\footnote{We use $\kappa > 0.01$, $\delta_{\text{bacc}} = 0.03$, $\delta_{\text{F1}} = 0.00$, and $\tau_{\text{dom}} = 0.95$, where \emph{maj} denotes the always-majority classifier.}—each targets a distinct failure mode:
\begin{itemize}
\item \textbf{Chance-corrected agreement:} Cohen's $\kappa > 0$.
\item \textbf{Imbalance-robust accuracy:} Balanced accuracy $> 1/K + \delta_{\text{bacc}}$.
\item \textbf{Macro-F1 dominance:} $F1_{\text{macro}} \geq F1^{(\text{maj})}_{\text{macro}} + \delta_{\text{F1}}$.
\item \textbf{Non-collapse checks:} At least two predicted classes; dominant predicted class fraction $\leq \tau_{\text{dom}}$.
\item \textbf{Evaluation setup:} $N = 50$ and $K \geq 2$.
\end{itemize}

Tasks failing to meet all criteria are \emph{skipped}, and only tasks passing all criteria are used in pretraining.

\paragraph{(ii) Example-level label-prediction consensus filter (warm-up only).}
For admitted tasks, we retain in the prompt only those evaluation examples where the RF prediction matches the ground truth (RF $=$ GT), discarding mismatched cases while keeping original labels and splits unchanged. This enforces teacher–student alignment without relabeling, yielding cleaner supervision in the earliest updates. After the warm-up stage, consensus filtering is discontinued, allowing the model to move beyond imitation and develop standalone in-context policies. 

After filtering, we reduce the evaluation set size from $N=50$ to $N_{conse}=20$. For the remaining usable examples, we resample within each label class to restore $N_{conse}=20$, aiming to match the proportions of many-shot demonstrations. This prevents collapse-to-majority under imbalance and avoids sub-optimal supervision.

\subsection{Token-efficient Prompting for In-Context ML}
\label{sec:prompting}
The application of many-shot in-context ML is often constrained by the fixed context length and high computational cost of long contexts~\citep{gardner2024largescale}. 
We alleviate these limitations through three design choices: \textbf{(i)} compact structuring of in-context examples, \textbf{(ii)} compression of numeric tokens, and \textbf{(iii)} sequence-level batching.

\subsubsection{Tabular encoding}
Instead of presenting demonstrations $S$ as scattered NL descriptions~\citep{hegselmann2023tabllm,gardner2024largescale}, we place many-shot examples in a concise table-style format. Building on recent findings that normal tabular structures can be seamlessly understood by popular LLMs~\citep{dong2024spreadsheetllm,sui2023tablemeetsllm}, our approach preserves column/field meaning while dramatically reducing tokens relative to full-sentence templates.

\begin{figure}[t]
  \centering
  \includegraphics[width=\columnwidth]{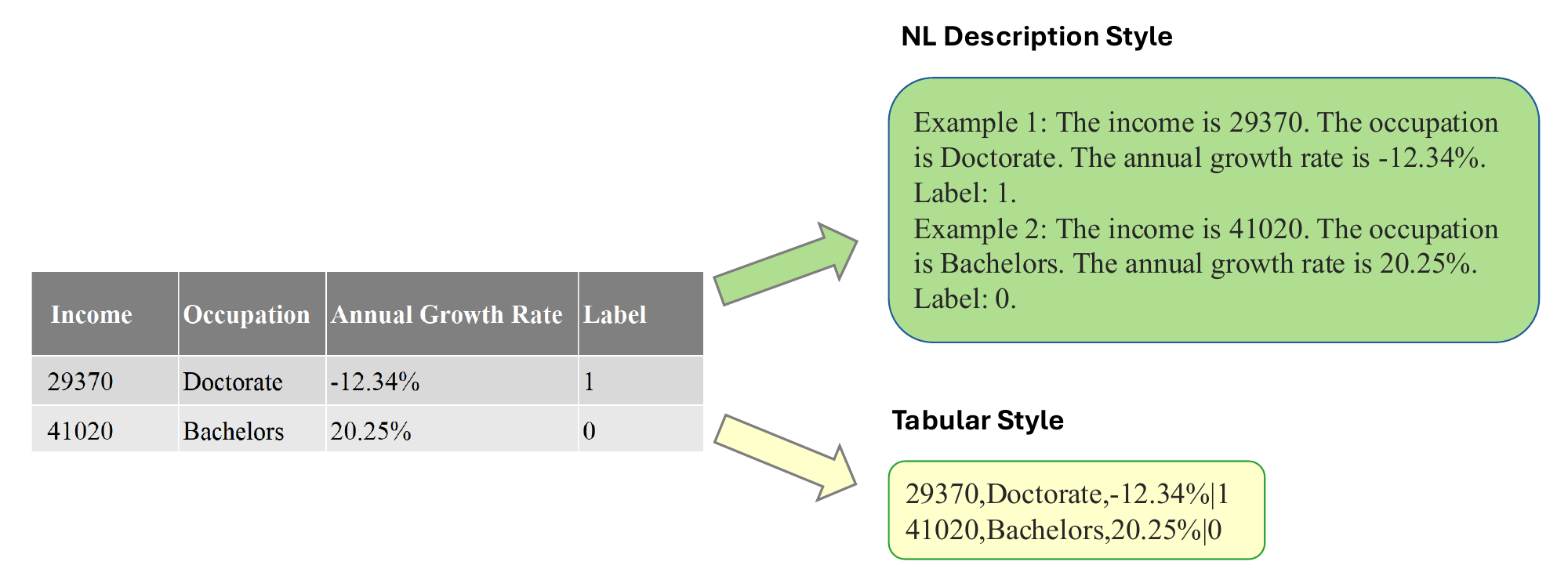}

  \caption{A comparison between the NL description style and tabular style of many-shot examples.}
  \vspace{-1em}
  \label{fig:tabular_style}
\end{figure}

The example in \ref{fig:tabular_style} contrasts an encoding table expressed as scattered NL descriptions, as in TabLLM \citep{hegselmann2023tabllm} and TabuLa \citep{gardner2024largescale}. with our compact tabular encoding, where commas separate features and ``|'' separates the label (the header is optional).  

Notably, while feature names and task descriptions may contain contextual semantics that could be readily exploited by LLMs, our approach \emph{avoids} using such information. This ensures a fair comparison with traditional ML methods and avoids data contamination and memorization~\citep{bordt2024elephants}. Moreover, the current tabular encoding jointly represents heterogeneous textual and numeric features, \emph{without requiring mandatory categorical bucketing or embedding extraction} for text. Toward future extensions of current tabular encoding to \emph{multimodal} features, HTML tables can be leveraged to support images, hyperlinks, and embedded objects.

\subsubsection{Compact integral-based number encoding}
We normalize all numbers of each numerical feature to non-negative integers in $[0,999]$ to eliminate the tokenization fragmentation caused by decimal points (``.'') and leading signs (``+''/``–''). The resulting values are then represented as plain text. This preserves ordinal structure yet maps each number from a sequence of scattered tokens to an integral single token under GPT's \texttt{cl100k\_base}. Beyond token savings, integer rendering avoids a frequent pitfall where decimals are compared text-wise (e.g., ``1.11'' \texttt{(1|.|11)} vs.\ ``1.9''  \texttt{(1|.|9} where 11 is bigger than 9 while 1.11 is smaller than 1.9)) rather than numerically~\citep{spithourakis2018numeracy,wallace2019numbers,singh2024tokenizationcounts,hfnumtokblog2024}.

\paragraph{Connection to $z$-norm.}
Recent work in tabular modeling, including TabICL~\citep{qu2025tabicl}, has adopted $z$-score normalization (mean--variance standardization) for each feature, which yields real-valued decimals. We also adopt this approach as the initial step before discretizing the normalized values into bounded integers in $[0,999]$.

\paragraph{Integer mapping.}
We map each normalized value $z$ to an integer $i\in[0,999]$ via
$i=\operatorname{clip}\!\big(\operatorname{round}(120z+500),\,0,\,999\big)$,
where $\operatorname{round}(x)=\lfloor x+0.5\rfloor$ and 
$\operatorname{clip}(x,a,b)=\min\{\max(x,a),\,b\}$.

Under this mapping, $z=0$ is mapped to $i=500$, and $z=\pm 4.166$--which covers $99.997\%$ of values under the standard normal distribution-- maps approximately to $i=0-999$, with out-of-band values clipped. This transformation yields \emph{[0,999]-norm}, which preserves numeric order while ensuring that most values are tokenized as a single token under \texttt{cl100k\_base}. Specifically, GPT (cl100k\_base) and LLaMA-3 vocabularies merge consecutive digits, commonly treating a sequence of three digits as a single token. In contrast, Qwen and LLaMA-2 vocabularies tokenize numbers at the level of individual digits, leading to much more fragmented representations. Below illustrates typical fragmentations of popular tokenizers for z-norm decimals and [0-999] non-negative integers.

\begin{tabular}{lll}
\multicolumn{3}{l}{\textbf{Z-norm (decimal form)}} \\[0.3em]

GPT (cl100k\_base), LLaMA-3: & $-0.1234$ & $\to$ 
  \tok{msOrange}{-}|\tok{msGreen}{0}|\tok{msYellow}{.}|\tok{msBlue}{123}|\tok{msGreen}{4} \\

Qwen, LLaMA-2: & $-0.1234$ & $\to$ 
  \tok{msOrange}{-}|\tok{msGreen}{0}|\tok{msYellow}{.}|\tok{msBlue}{1}|\tok{msYellow}{2}|\tok{msGreen}{3}|\tok{msBlue}{4} \\[1em]

\multicolumn{3}{l}{\textbf{[0,999]-norm (integer form)}} \\[0.3em]

GPT (cl100k\_base), LLaMA-3: & $486$ & $\to$ 
  \tok{msBlue}{486} \\

Qwen, LLaMA-2: & $486$ & $\to$ 
  \tok{msBlue}{4}|\tok{msYellow}{8}|\tok{msGreen}{6} \\
\end{tabular}

\subsubsection{Sequence-level batch-prediction}
\label{sec:seq_batch_pred}
As shown in Figure~\ref{fig:cover}, we \emph{pack} $N$ query rows into a single sequence and decode all predictions in one forward pass, increasing the \emph{effective} batch size at the sequence level~\citep{lin2023batchprompt,cheng2023batch}. 
Let the sequence consist of a shared header $H$ (task instruction, schema) and in-context demonstrations $S$, followed by $N$ test queries $\{x^{(i)}\}_{i=1}^{N}$ and an output section for predictions $\{y^{(i)}\}_{i=1}^{N}$:
\[
\underbrace{H \,\Vert\, S}_{\text{instruction + schema + $M$ shots}}
\,\Vert\,
\underbrace{x^{(1)} \,\Vert\, \cdots \,\Vert\, x^{(N)}}_{\text{batched queries}}
\;\;\Longrightarrow\;\;
\underbrace{y^{(1)} \,\Vert\, \cdots \,\Vert\, y^{(N)}}_{\text{batched predictions}}.
\]
During training, \textsc{MachineLearningLM} remains purely autoregressive and is optimized with the standard next-token objective (Section~\ref{sec:objective}). Increasing $N$ accelerates and stabilizes continued pretraining. At inference, the model predicts $N$ samples in a single pass, amortizing instruction and context overhead (Section~\ref{compression}).

\paragraph{Stability and reliability.}
A big $N$ not only consumes the token budget that would otherwise be allocated to demonstrations, but also increases instruction-following errors (e.g., cross-item interference and run-on/non-terminated generations), particularly for smaller open-source models. We therefore set $N\!=\!50$ by default. In addition, we randomize the within-sequence order of $\{x^{(i)}\}$ to reduce position bias and improve exchangeability of the batched items.\footnote{This simple permutation also helps mitigate long-context position effects observed in batch-prediction.} Importantly, \texttt{id} should be 0-indexed; 1-based indexing often destabilizes instruction following and leads to incorrect IDs.

\subsubsection{Compression and Amortization Ratio}
\label{compression}

\paragraph{Token-cost model.}

We model the amortized token cost per predicted label as follows.

Let $M$ denote the number of many-shot demonstrations (with labels) and $N$ be the number of query examples to be inferred (with IDs). Each row is dominated by its $d$ features, so we represent the per-row token cost by $R$, without distinguishing labels and IDs. The instruction/schema header incurs a token cost of $H$, which is negligible compared to $R(M+N)$ and can be omitted. The resulting amortized token cost per predicted label is therefore
\[
C \ = \frac{H+R\,(M+N)}{N} \approx\; \frac{R\,(M+N)}{N}.
\]

\paragraph{Compression and amortization ratio.}
The overall compression ratio is defined as the product of three saving factors: (A) \emph{tabular structure}, (B) \emph{number normalization}, and (C) \emph{batch inference}

\subparagraph{(a) Tabular structure (NL--Decimal $\to$ Tabular--Decimal).}
We replace scattered NL sentences (e.g., ``\texttt{the first feature is 0.1234}) '' with compact, comma-delimited rows, and we apply this format throughout Section~\ref{compression}. This will substantially reduce the per-row cost $R$.
Counting words, spaces (absorbed into word tokens), numbers, and delimiters, we obtain:

\[
\begin{array}{lccc}
\toprule
\textbf{Model / Tokenizer} & R_{\text{NL,dec}} & R_{\text{Tab,dec}} & \textbf{Ratio} \\
\midrule
\text{GPT(cl100k\_base), LLaMA-3} & 10 & 5 & \mathbf{2.0}\times \\
\text{Qwen, LLaMA-2}              & 12 & 7 & \mathbf{1.71}\times \\
\bottomrule
\end{array}
\]

\subparagraph{(b) Number normalization.}
Mapping normalized decimals to the integer range $[0,999]$ preserves the relative rank of decimals while collapsing numbers to a single token under \texttt{cl100k\_base} (and 2–3 tokens under Qwen's tokenizer). 

\noindent\emph{Without delimiters.} 
For example, ``0.1234'' requires 4 tokens, whereas ``234'' requires only 1;similarly, ``--0.1234'' requires 5 tokens, compared to \ 1 for ``234''. 
Assuming a balanced distribution of positive/negative numbers, the expected token length is $(0.5\times 4 + 0.5\times 5)=4.5$ before normalization and $1$ after. 
Thus, the expected per-number reduction under \texttt{cl100k\_base} is $\mathbf{4.5}\times$. 
In Qwen's tokenizer, positives use 6 tokens vs.\ 3 after, negatives 7 vs.\ 3 after; the expected length is $(0.5\times 6 + 0.5\times 7)=6.5$ before and $3$ after, giving $6.5/3 \approx \mathbf{2.17}\times$. 

\noindent\emph{With delimiters.} 
In practice, each number will be followed by a delimiter (a single-token comma), which increases numerator and denominator lengths by $+1$. 
Hence for \texttt{cl100k\_base}, this yields an expected length of $5.5$ before normalization versus $2$ after, giving $5.5/2 = \mathbf{2.75}\times$. 
For Qwen’s tokenizer, the corresponding expectation is $7.5$ before versus $4$ after, resulting in $7.5/4 \approx \mathbf{1.88}\times$.

\noindent These expected compression ratios propagate to corpus-level savings:

\[
\begin{array}{lcc}
\toprule
\textbf{Model / Tokenizer} & \textbf{w/o delimiters} & \textbf{w/ delimiters} \\
\midrule
\text{GPT(cl100k\_base), LLaMA-3} & \mathbf{4.5}\times & \mathbf{2.75}\times \\
\text{Qwen, LLaMA-2}              & \mathbf{2.17}\times & \mathbf{1.88}\times \\
\bottomrule
\end{array}
\]

\subparagraph{(c) Batch inference.}
Given $C(N)=\tfrac{R(M+N)}{N}$, increasing $N$ amortizes the many-shot block.

\noindent General form:
\[
\frac{C(1)}{C(N)}=\frac{N(M+1)}{M+N}\quad(\text{independent of tokenizer since $R$ cancels}).
\]
\noindent Derivation for $M{=}1024$, from $N{=}1$ to $N{=}50$:
\[
\frac{C(N=1)}{C(N=50)}=\frac{50\times(1024+1)}{(1024+50)}=47.7\times.
\]

\paragraph{Overall compression and amortization.} 
Since Stages (A), (B), and (C) are independent, their effects compound multiplicatively.
From a compression perspective, (A) structural formatting and (B) number normalization compound multiplicatively to shrink the per-row token cost $R$, which then will allow the LLM context to accommodate more many-shot demonstrations. 

\noindent Hence \emph{Compression (a$\times$b):}
\[
\text{GPT(cl100k\_base), LLaMA 3: }  2.0 \times 2.75 \;\approx\; \boxed{5.5\times},
\qquad
\text{Qwen, LLaMA 2: } 1.71 \times 1.88 \;\approx\; \boxed{3.2\times}.
\]

\noindent\emph{Amortization (c, $n{=}50$):}
\[
\boxed{47.7\times}
\]

Notably, the above estimate assumes that the prompt header $H$ contributes negligibly to the overall computation. Beyond this estimation, we leverage our SCM-based synthetic pretraining corpus to evaluate the amortized token cost per inference sample using both our encoding and TabuLa~\citep{gardner2024largescale}’s. Under this setup, we observe a 136× improvement using Qwen's tokenizer.

\subsection{Order-robust, confidence-aware self-consistency at inference time}
\label{sec:order_robust}

Our approach is similar to self-consistency proposed by \citep{wang2022self}, but modifies both the source of diversity and the aggregation method. Instead of generating diverse reasoning paths from identical prompts via stochastic sampling, we create diversity by shuffling in-context demonstrations or features across prompt variants. We then aggregate the model's responses using weighted majority voting \citep{littlestone1994weighted} to select the most consistent prediction.

More specifically, given a prompt with $M$ in-context demonstrations, we generate $V$ shuffled variants $\{P_0, P_1, \ldots, P_{V-1}\}$ and obtain the model's next-token probabilities for each variant via parsing responses of LLMs. For each query, we extract the corresponding probability of the next token for each possible label $y_j \in \{y_1, y_2, \ldots, y_k\}$ from the model's vocabulary distribution. For prompt variant $P_i$, let $p_i(y_j)$ denote the probability assigned by the LLM to the token representing label $y_j$ as the next token in the sequence. We compute the aggregate probability for each label by summing across all prompt variants:

\begin{equation}
\tilde{p}(y_j) = \sum_{i=0}^{V-1} p_i(y_j)
\end{equation}

The final prediction is then determined by selecting the label with the highest aggregate probability: $\hat{y} = \arg\max_{y_j} \tilde{p}(y_j)$.
We set $V=5$ in the following experiments—far fewer than the 32 used by TabICL and TabPFN.

\section{Experiments}

\subsection{Preliminary}

\subsubsection{Evaluation Datasets} 

We evaluate on the TALENT benchmark~\citep{ye2024closer}, which comprises 200 classification tasks, several of which include both tabular and textual features. For the main study, we select 32 datasets via domain clustering and sampling, and add 86 more as an extended set. We exclude one dataset whose combined numerical and textual fields exceeded our context-length budget. All tasks are listed in the Appendix. Each dataset is split 80/20 into train/test, and we report accuracy (ACC).

\subsubsection{Baselines}\label{sec:baselines}

\paragraph{Tree-based learners.}
We include classical tree ensembles as strong tabular baselines: Random Forest (RF)~\citep{breiman2001randomforests}, LightGBM~\citep{ke2017lightgbm}, XGBoost~\citep{chen2016xgboost}, and CatBoost~\citep{prokhorenkova2018catboost}. 
In our main comparisons, we report \emph{Random Forest} because it is stable and robust from few-shot to many-shot regimes and easy to reproduce at scale. 
We use the same hyperparameters as the RF teacher in the LLM experiments. For Random Forest, we set \texttt{n\_estimators=30}, fix \texttt{random\_state} for determinism, and use \texttt{n\_jobs=8} (we already parallelize across processes).

\paragraph{Instance-based learner.}
Recent work by~\citet{agarwal2024manyshot} suggests that LLMs can behave similarly to local neural networks when learning linear high-dimensional functions with numerical inputs. Motivated by this observation, we include $k$-nearest neighbors ($k$NN)~\citep{cover1967nearest} as a simple instance-based baseline. Specifically, we use \texttt{n\_neighbors=8}, \texttt{weights= ``distance''}, and the Minkowski distance with \texttt{p=2}.

\paragraph{Tabular ICL models.}
We compare to \textbf{TabICL}~\citep{qu2025tabicl}, a recent in-context learners for tabular data. 
We run their public ICL-style inference (no gradient updates), which is faster to evaluate and aligns with our in-context setup. We also compare against \textbf{TabuLa-8B}~\citep{gardner2024largescale}, an 8B LLM fine-tuned specifically for tabular prediction.

\paragraph{General-purpose LLMs.}
For closed-source LLMs, we evaluate \textbf{GPT-5-mini} (latest public snapshot) and the reasoning-oriented \textbf{o3-mini}~\citep{openai-o3-mini-changelog,openai-o3-o4mini-systemcard}. 
For open-source LLMs, we use \textbf{Qwen-2.5-7B-Instruct}~\citep{bai2023qwen} as a strong lightweight baseline. 
\textsc{MachineLearningLM} is also built on Qwen-2.5-7B-Instruct via continued pretraining, enabling a direct apples-to-apples comparison under identical prompting.

\subsubsection{Continued pretraining setting introduction} 

We employ the task generator described in Section~\ref{synthesis_data} to construct our synthetic pretraining corpus. Our dataset comprises 3 million synthetic tasks, each containing no more than 1,024 samples with feature dimensions ranging from 5 to 50 and up to 10 class labels. Owing to the 32k-token context limit during training, we truncate the number of in-context examples per task during continued pretraining.

We build upon Qwen-7B-Instruct~\citep{bai2023qwen} as our backbone model, which utilizes rotary positional embeddings. We adopt a two-stage continual pretraining approach, implemented using LoRA~\citep{hu2022lora} with rank 8 over attn+MLP and the Adam optimizer~\citep{kingma2014adam}, within the LLaMAFactory framework~\citep{zheng2024llamafactory}. Training was distributed across 5 nodes with 40 A100 GPUs, achieving a throughput of 300 tokens per second (100s per iteration, with each iteration processing $\sim$30k tokens).
We set lr\_scheduler\_type as cosine to use the cosine learning rate scheduler throughout training. In the first (warm-up) stage, we set the learning rate to $1 \times 10^{-5}$ and trained on approximately 1 million tasks over 100 hours. The second stage resumed from the Stage 1 checkpoint with a reduced learning rate of $1 \times 10^{-6}$, continuing on about 2 million tasks for 200 hours.
We set the batch size per GPU to 1, and updated model parameters every 8 steps. With 40 GPUs, this corresponds to one parameter update per 320 samples.

\subsubsection{Test time setting} 

For ICL evaluation, we partition test data into chunks of size $N$, with each chunk serving as a test batch. For each test chunk, we randomly sample $M$ training examples as in-context demonstrations, and we use different sets of
$M$ examples for each batch to ensure the results are not biased by a particular fixed set. When generating prompt variants as described in Section \ref{sec:order_robust}, we shuffle only the order of the $M$ in-context examples while keeping the test examples and their order fixed.
For fair comparison with traditional ML models, we randomly sample $M$ training examples for model training and evaluate on the complete test set, as detailed in our public code. 

Importantly, although the header row in tabular data could be used to incorporate feature names/descriptions that LLMs could easily leverage~\citep{bordt2024elephants}, our method \emph{intentionally excludes such information} during training and inference. This design choice promotes a fair comparison with conventional machine learning approaches and mitigates risks of data leakage and memorization~\citep{bordt2024elephants}, while still delivering competitive performance. Additionally, our current tabular encoding strategy jointly models both textual and numerical features, without relying on categorical bucketing or explicit text embedding extraction. The number of $S$ samples for voting is 5 due to computational constraints, while TabICL and TabPFN use 32.

\subsection{Experiment Results}

\subsubsection{\textsc{MachineLearningLM} vs.\ vanilla LLMs}
As Table~\ref{tab:shots_curve} and Table~\ref{tab:shots_curve_tasks_32_512_merged} show, pretraining solely on synthetic tabular tasks yields \emph{large, consistent} gains over vanilla LLMs. Averaged across tasks, \textsc{MachineLearningLM} improves absolute accuracy of the backbone Qwen-2.5-7B-Instruct model by \textbf{$\sim$15\%} (Table~\ref{tab:shots_curve}), with about {$\sim$50\% of tasks seeing 15\% improvements across finance, biology, vision, speech, robotics, statistics, and healthcare domains (Table~\ref{tab:shots_curve_tasks_32_512_merged}); virtually no task exhibits a marked degradation (Table~\ref{tab:shots_curve_tasks_32_512_merged}). At high shot counts (128–1,024), \textsc{MachineLearningLM} \emph{surpasses GPT-5-mini} by \textbf{$\sim$12\%} and o3-mini by \textbf{$\sim$16\%} on average (cf.\ Figure~\ref{fig:cover}c and Table~\ref{tab:shots_curve}). More results on extended tasks are presented in Appendix~\ref{extend_eval}.

\subsubsection{Many-shot scaling}
\textsc{MachineLearningLM} displays a \emph{striking many-shot scaling law}: accuracy increases monotonically as we raise the in-prompt demonstrations from \(2^3\) to \(2^{10}\) (within a 131k-token inference budget, \textbf{4$\times$} the 32k-token budget in the pretraining phase), with steady gains across domains (Figure~\ref{fig:cover}c; Table~\ref{tab:shots_curve}). While our model can improve accuracy over 15\% from 8\ to 512 shots, vanilla LLMs exhibit limited scaling: o3-mini improves by only \(\sim\)1.8\% (and even \emph{declines} from 64 to 512 shots), while GPT-5-mini gains merely \(\sim\)4.7\%. This indicates substantially higher sample efficiency of our approach.

\subsubsection{\textsc{MachineLearningLM} vs.\ TabuLa-8B}
Thanks to the \emph{token-efficient prompting}, \textsc{MachineLearningLM} supports an order of magnitude more demonstrations per prompt than TabuLa-8B (Table~\ref{tab:shots_curve_tasks_32_512_merged}), which is typically capped by an 8k context (often \(\le\)20–32 shots). In contrast, we scale stably to \textbf{1,024} shots for most tasks while \emph{maintaining} significant many-shot gains, as shown in Appendix~\ref{extend_shot_eval}. In the few-shot regime (e.g., 32-shot), \textsc{MachineLearningLM} also outperforms TabuLa-8B on average—even though TabuLa explicitly exploits feature names— with particularly large gaps on certain datasets (e.g., a \(\sim\)14.5\% shortfall for \textit{vehicle} reported by TabuLa).

\providecommand{\negcolor}[1]{#1}
\definecolor{ml-lime}{HTML}{C9DA2A}

\renewcommand{\negcolor}[1]{%
  \begingroup
  \dimen0=#1pt\relax
  \ifdim\dimen0<1.5pt \dimen0=100\dimen0\fi
  \ifdim\dimen0<40pt
    \cellcolor{ml-lime!5!white}{#1}
  \else\ifdim\dimen0<45pt
    \cellcolor{ml-lime!10!white}{#1}
  \else\ifdim\dimen0<50pt
    \cellcolor{ml-lime!18!white}{#1}
  \else\ifdim\dimen0<55pt
    \cellcolor{ml-lime!26!white}{#1}
  \else\ifdim\dimen0<60pt
    \cellcolor{ml-lime!34!white}{#1}
  \else\ifdim\dimen0<65pt
    \cellcolor{ml-lime!42!white}{#1}
  \else\ifdim\dimen0<70pt
    \cellcolor{ml-lime!50!white}{#1}
  \else\ifdim\dimen0<75pt
    \cellcolor{ml-lime!58!white}{#1}
  \else\ifdim\dimen0<80pt
    \cellcolor{ml-lime!66!white}{#1}
  \else\ifdim\dimen0<86pt
    \cellcolor{ml-lime!80!white}{#1}
  \else
    \cellcolor{ml-lime!85!white}{#1}
  \fi\fi\fi\fi\fi\fi\fi\fi\fi\fi
  \endgroup
}

\begin{table}[t]
\centering
\small
\setlength{\tabcolsep}{6pt}
\begin{threeparttable}
\caption{Test accuracy (\%) across number of shots $M{=}8$–$512$. Notably, as shown in Figure~\ref{fig:heatmap}, a large fraction of training tasks exceed the 32k-token limit when both the number of shots $M$ and number of features $K$ are big; nevertheless, the model exhibits out-of-distribution generalization to 131k-token contexts. Abbreviations: \;Qwen-7B$=$Qwen-2.5-7B-Instruct;\;Ours$=$\textsc{MachineLearningLM} (base: Qwen-2.5-7B-Instruct). }
\label{tab:shots_curve}
\begin{tabular}{l cc ccccc}
\toprule
 & \multicolumn{2}{c}{\textbf{Not in-context learning}} & \multicolumn{5}{c}{\textbf{In-context learning}} \\
\cmidrule(lr){2-3}\cmidrule(lr){4-8}
\textbf{Number of shots} & \textbf{KNN} & \textbf{Random Forest} & \textbf{TabICL} & \textbf{GPT-5 mini} & \textbf{o3-mini} & \textbf{Qwen-7B} & \textbf{Ours} \\
\midrule
8     & \negcolor{55.3} & \negcolor{59.1} & \negcolor{57.4} & \negcolor{57.8} & \negcolor{57.0} & \negcolor{51.8} & \negcolor{58.4} \\
16    & \negcolor{60.1} & \negcolor{63.5} & \negcolor{64.0} & \negcolor{60.0} & \negcolor{58.6} & \negcolor{53.9} & \negcolor{63.1} \\
32    & \negcolor{63.4} & \negcolor{68.0} & \negcolor{68.6} & \negcolor{60.2} & \negcolor{59.0} & \negcolor{55.6} & \negcolor{66.7} \\
64    & \negcolor{65.9} & \negcolor{71.4} & \negcolor{72.6} & \negcolor{60.4} & \negcolor{59.6} & \negcolor{57.2} & \negcolor{70.0} \\
128   & \negcolor{68.1} & \negcolor{74.3} & \negcolor{76.0} & \negcolor{61.0} & \negcolor{58.4} & \negcolor{58.6} & \negcolor{72.0} \\
256   & \negcolor{69.5} & \negcolor{76.1} & \negcolor{79.1} & \negcolor{61.7} & \negcolor{58.8} & \negcolor{59.3} & \negcolor{74.3} \\
512   & \negcolor{71.1} & \negcolor{77.7} & \negcolor{80.9} & \negcolor{62.5} & \negcolor{58.8} & \negcolor{60.1} & \negcolor{75.3} \\
\bottomrule
\end{tabular}
\end{threeparttable}
\end{table}

\subsubsection{Competitive many-shot performance vs.\ state-of-the-art tabular-specific methods}

Without any task-specific training, \textsc{MachineLearningLM} reaches \emph{random-forest–level} accuracy across $M{=}8$–$512$ shots—typically within \textbf{2\%} \emph{relative} (Table~\ref{tab:shots_curve})—and, even at 512 shots, outperforms random forest on roughly \textbf{30\%} of tasks (Table~\ref{tab:shots_curve_tasks_32_512_merged}). It also \emph{clearly} surpasses simple instance-based learners (e.g., kNN) by \textbf{>\,4\%} \emph{relative} on average, indicating robust numerical modeling (Table~\ref{tab:shots_curve}; Table~\ref{tab:shots_curve_tasks_32_512_merged}).

\begin{table}[]
\vspace{-1.6em}
\centering
\fontsize{8pt}{9pt}\selectfont
\setlength{\tabcolsep}{6pt}
\begin{threeparttable}
\caption{Per-task test accuracy (\%) with $M\in\{32,512\}$ shots. Abbreviations: Tabula $=$ TabuLa-8B; Qwen $=$ Qwen-2.5-7B-Instruct; Ours $=$ \textsc{MachineLearningLM} (base: Qwen-2.5-7B-Instruct). To reduce memorization risks~\citep{bordt2024elephants}, all methods drop feature names/descriptions and encode only values except for TabuLa-8B, because we adopt the paper’s results on the overlapping datasets, using the setting that keeps the original header names. ``EL'' indicates the prompt exceeds TabuLa-8B’s token limit; ``NA'' indicates the dataset is not included in TabuLa-8B’s evaluation set.}
\label{tab:shots_curve_tasks_32_512_merged}
\begin{tabular}{l r c c c c c c c c}
\toprule
\textbf{Task} & \textbf{\# Shots} & \textbf{KNN} & \textbf{RF} & \textbf{TabICL} & \textbf{Tabula} & \textbf{GPT-5-mini} & \textbf{o3-mini} & \textbf{Qwen} & \textbf{Ours} \\
\midrule
Bank & 32  & \negcolor{87.3} & \negcolor{87.6} & \negcolor{88.3} & \negcolor{84.4} & \negcolor{85.4} & \negcolor{85.3} & \negcolor{87.2} & \negcolor{87.3} \\
Bank & 512 & \negcolor{87.6} & \negcolor{89.0} & \negcolor{89.4} & EL & \negcolor{87.8} & \negcolor{86.7} & \negcolor{88.1} & \negcolor{88.7} \\
BLE\_RSSI\_Indoor\_Loc. & 32  & \negcolor{33.7} & \negcolor{61.5} & \negcolor{61.7} & NA & \negcolor{59.4} & \negcolor{61.2} & \negcolor{44.7} & \negcolor{63.7} \\
BLE\_RSSI\_Indoor\_Loc. & 512 & \negcolor{35.1} & \negcolor{69.2} & \negcolor{73.5} & EL & \negcolor{61.4} & \negcolor{58.4} & \negcolor{52.7} & \negcolor{70.3} \\
Churn & 32  & \negcolor{84.6} & \negcolor{86.5} & \negcolor{86.6} & \negcolor{91.4} & \negcolor{82.1} & \negcolor{79.1} & \negcolor{85.4} & \negcolor{86.7} \\
Churn & 512 & \negcolor{86.6} & \negcolor{91.1} & \negcolor{92.0} & EL & \negcolor{86.1} & \negcolor{83.6} & \negcolor{86.2} & \negcolor{89.5} \\
CMC & 32  & \negcolor{45.1} & \negcolor{45.8} & \negcolor{46.1} & \negcolor{35.2} & \negcolor{42.7} & \negcolor{41.4} & \negcolor{37.6} & \negcolor{44.1} \\
CMC & 512 & \negcolor{52.2} & \negcolor{50.8} & \negcolor{55.6} & EL & \negcolor{41.0} & \negcolor{42.7} & \negcolor{50.2} & \negcolor{50.2} \\
Contaminant\_10\_0GHz & 32  & \negcolor{69.4} & \negcolor{72.5} & \negcolor{73.3} & NA & \negcolor{60.8} & \negcolor{53.3} & \negcolor{55.6} & \negcolor{71.7} \\
Contaminant\_10\_0GHz & 512 & \negcolor{78.8} & \negcolor{85.6} & \negcolor{92.1} & EL & \negcolor{60.4} & \negcolor{49.4} & \negcolor{62.7} & \negcolor{82.3} \\
Contaminant\_9\_5GHz & 32  & \negcolor{71.9} & \negcolor{71.0} & \negcolor{71.0} & NA & \negcolor{55.0} & \negcolor{53.8} & \negcolor{53.5} & \negcolor{73.8} \\
Contaminant\_9\_5GHz & 512 & \negcolor{79.4} & \negcolor{84.6} & \negcolor{89.2} & EL & \negcolor{53.1} & \negcolor{48.5} & \negcolor{59.4} & \negcolor{80.4} \\
Credit\_g & 32  & \negcolor{64.0} & \negcolor{69.0} & \negcolor{72.0} & \negcolor{70.3} & \negcolor{70.0} & \negcolor{63.5} & \negcolor{58.5} & \negcolor{71.5} \\
Credit\_g & 512 & \negcolor{70.0} & \negcolor{76.0} & \negcolor{76.5} & EL & \negcolor{71.5} & \negcolor{68.0} & \negcolor{71.0} & \negcolor{71.5} \\
FICO\_HELOC\_Cleaned & 32  & \negcolor{59.0} & \negcolor{64.7} & \negcolor{68.3} & NA & \negcolor{54.6} & \negcolor{57.6} & \negcolor{54.0} & \negcolor{63.1} \\
FICO\_HELOC\_Cleaned & 512 & \negcolor{64.3} & \negcolor{71.5} & \negcolor{71.3} & EL & \negcolor{54.7} & \negcolor{52.6} & \negcolor{52.6} & \negcolor{69.0} \\
FOREX\_AUDCHF\_Day & 32  & \negcolor{50.7} & \negcolor{52.3} & \negcolor{51.2} & EL & \negcolor{48.8} & \negcolor{47.9} & \negcolor{49.9} & \negcolor{51.2} \\
FOREX\_AUDCHF\_Day & 512 & \negcolor{44.4} & \negcolor{52.6} & \negcolor{71.1} & EL & \negcolor{51.2} & \negcolor{49.0} & \negcolor{54.5} & \negcolor{49.6} \\
FOREX\_AUDJPY\_Day & 32  & \negcolor{45.5} & \negcolor{46.6} & \negcolor{49.0} & NA & \negcolor{46.6} & \negcolor{46.3} & \negcolor{45.8} & \negcolor{46.9} \\
FOREX\_AUDJPY\_Day & 512 & \negcolor{57.8} & \negcolor{61.3} & \negcolor{72.8} & EL & \negcolor{54.2} & \negcolor{47.4} & \negcolor{49.3} & \negcolor{62.4} \\
GAMETES\_Heterog. & 32  & \negcolor{47.5} & \negcolor{51.3} & \negcolor{46.3} & NA & \negcolor{52.2} & \negcolor{48.8} & \negcolor{48.1} & \negcolor{49.1} \\
GAMETES\_Heterog. & 512 & \negcolor{57.8} & \negcolor{59.7} & \negcolor{71.3} & EL & \negcolor{52.2} & \negcolor{49.1} & \negcolor{49.7} & \negcolor{52.2} \\
HELOC & 32  & \negcolor{59.8} & \negcolor{66.0} & \negcolor{58.7} & EL & \negcolor{49.5} & \negcolor{51.8} & \negcolor{52.1} & \negcolor{62.5} \\
HELOC & 512 & \negcolor{65.6} & \negcolor{71.3} & \negcolor{72.3} & EL & \negcolor{51.4} & \negcolor{51.0} & \negcolor{53.0} & \negcolor{70.8} \\
KC1 & 32  & \negcolor{81.8} & \negcolor{79.9} & \negcolor{78.4} & \negcolor{82.0} & \negcolor{78.9} & \negcolor{75.8} & \negcolor{81.8} & \negcolor{78.4} \\
KC1 & 512 & \negcolor{82.5} & \negcolor{82.7} & \negcolor{84.4} & EL & \negcolor{81.9} & \negcolor{78.7} & \negcolor{83.4} & \negcolor{83.9} \\
LED24 & 32  & \negcolor{30.8} & \negcolor{41.9} & \negcolor{46.4} & NA & \negcolor{8.3} & \negcolor{14.5} & \negcolor{13.1} & \negcolor{15.3} \\
LED24 & 512 & \negcolor{55.8} & \negcolor{69.4} & \negcolor{71.7} & EL & \negcolor{9.1} & \negcolor{12.3} & \negcolor{11.7} & \negcolor{48.4} \\
LED7 & 32  & \negcolor{55.9} & \negcolor{57.7} & \negcolor{58.8} & EL & \negcolor{55.9} & \negcolor{55.5} & \negcolor{32.8} & \negcolor{58.4} \\
LED7 & 512 & \negcolor{68.8} & \negcolor{68.8} & \negcolor{70.3} & EL & \negcolor{66.6} & \negcolor{57.3} & \negcolor{48.4} & \negcolor{66.9} \\
Maternal\_Health\_Risk & 32  & \negcolor{45.8} & \negcolor{52.7} & \negcolor{60.1} & NA & \negcolor{41.4} & \negcolor{52.7} & \negcolor{45.3} & \negcolor{50.7} \\
Maternal\_Health\_Risk & 512 & \negcolor{74.4} & \negcolor{78.3} & \negcolor{80.0} & EL & \negcolor{73.9} & \negcolor{63.0} & \negcolor{59.6} & \negcolor{78.3} \\
PC1 & 32  & \negcolor{92.3} & \negcolor{92.3} & \negcolor{90.1} & \negcolor{89.8} & \negcolor{88.7} & \negcolor{88.3} & \negcolor{85.1} & \negcolor{88.7} \\
PC1 & 512 & \negcolor{91.9} & \negcolor{93.7} & \negcolor{94.1} & EL & \negcolor{93.3} & \negcolor{89.2} & \negcolor{93.2} & \negcolor{93.2} \\
Phoneme & 32  & \negcolor{74.4} & \negcolor{75.3} & \negcolor{79.9} & \negcolor{73.4} & \negcolor{71.5} & \negcolor{71.9} & \negcolor{67.0} & \negcolor{73.5} \\
Phoneme & 512 & \negcolor{83.0} & \negcolor{84.9} & \negcolor{84.6} & EL & \negcolor{69.6} & \negcolor{64.3} & \negcolor{68.7} & \negcolor{82.0} \\
Pima\_Indians\_Diabetes & 32  & \negcolor{64.9} & \negcolor{74.7} & \negcolor{66.9} & \negcolor{70.3} & \negcolor{69.5} & \negcolor{61.0} & \negcolor{65.6} & \negcolor{71.4} \\
Pima\_Indians\_Diabetes & 512 & \negcolor{68.8} & \negcolor{75.3} & \negcolor{74.7} & EL & \negcolor{62.3} & \negcolor{68.2} & \negcolor{70.8} & \negcolor{75.3} \\
RingNorm & 32  & \negcolor{53.8} & \negcolor{79.6} & \negcolor{84.2} & NA & \negcolor{66.2} & \negcolor{75.9} & \negcolor{69.2} & \negcolor{93.2} \\
RingNorm & 512 & \negcolor{58.6} & \negcolor{93.5} & \negcolor{97.0} & EL & \negcolor{74.5} & \negcolor{75.3} & \negcolor{69.3} & \negcolor{96.0} \\
RL & 32  & \negcolor{56.6} & \negcolor{58.5} & \negcolor{55.8} & NA & \negcolor{53.5} & \negcolor{51.8} & \negcolor{53.9} & \negcolor{57.1} \\
RL & 512 & \negcolor{60.7} & \negcolor{66.7} & \negcolor{68.5} & EL & \negcolor{52.7} & \negcolor{54.8} & \negcolor{51.8} & \negcolor{64.1} \\
Segment & 32  & \negcolor{63.6} & \negcolor{68.2} & \negcolor{61.0} & EL & \negcolor{60.4} & \negcolor{51.5} & \negcolor{32.0} & \negcolor{64.7} \\
Segment & 512 & \negcolor{88.1} & \negcolor{90.9} & \negcolor{93.5} & EL & \negcolor{68.6} & \negcolor{52.0} & \negcolor{52.2} & \negcolor{87.2} \\
Seismic\_Bumps & 32  & \negcolor{93.2} & \negcolor{92.7} & \negcolor{93.0} & NA & \negcolor{91.3} & \negcolor{86.3} & \negcolor{92.5} & \negcolor{90.3} \\
Seismic\_Bumps & 512 & \negcolor{92.8} & \negcolor{92.8} & \negcolor{93.0} & EL & \negcolor{90.7} & \negcolor{86.5} & \negcolor{93.0} & \negcolor{90.7} \\
Statlog & 32  & \negcolor{61.0} & \negcolor{65.0} & \negcolor{64.0} & NA & \negcolor{60.0} & \negcolor{61.0} & \negcolor{57.5} & \negcolor{63.0} \\
Statlog & 512 & \negcolor{67.0} & \negcolor{74.0} & \negcolor{72.5} & EL & \negcolor{58.0} & \negcolor{63.0} & \negcolor{63.0} & \negcolor{67.0} \\
Thyroid & 32  & \negcolor{92.6} & \negcolor{92.8} & \negcolor{93.3} & NA & \negcolor{92.3} & \negcolor{91.2} & \negcolor{92.6} & \negcolor{92.8} \\
Thyroid & 512 & \negcolor{94.0} & \negcolor{98.0} & \negcolor{98.8} & EL & \negcolor{92.6} & \negcolor{92.5} & \negcolor{92.6} & \negcolor{95.0} \\
TwoNorm & 32  & \negcolor{94.4} & \negcolor{88.8} & \negcolor{95.9} & NA & \negcolor{88.2} & \negcolor{96.0} & \negcolor{62.9} & \negcolor{95.0} \\
TwoNorm & 512 & \negcolor{97.0} & \negcolor{95.9} & \negcolor{97.6} & EL & \negcolor{87.6} & \negcolor{93.1} & \negcolor{74.9} & \negcolor{97.5} \\
Vehicle & 32  & \negcolor{52.3} & \negcolor{60.0} & \negcolor{70.0} & \negcolor{48.4} & \negcolor{47.1} & \negcolor{24.7} & \negcolor{32.9} & \negcolor{62.9} \\
Vehicle & 512 & \negcolor{64.7} & \negcolor{74.7} & \negcolor{84.7} & EL & \negcolor{41.2} & \negcolor{27.1} & \negcolor{32.4} & \negcolor{72.9} \\
WallRobot\_Navigation & 32  & \negcolor{51.4} & \negcolor{74.2} & \negcolor{70.1} & NA & \negcolor{43.0} & \negcolor{35.8} & \negcolor{42.0} & \negcolor{60.1} \\
WallRobot\_Navigation & 512 & \negcolor{73.4} & \negcolor{96.7} & \negcolor{94.1} & EL & \negcolor{48.9} & \negcolor{35.8} & \negcolor{45.2} & \negcolor{83.3} \\
Waveform & 32  & \negcolor{74.1} & \negcolor{72.7} & \negcolor{76.0} & NA & \negcolor{44.2} & \negcolor{40.5} & \negcolor{38.6} & \negcolor{72.5} \\
Waveform & 512 & \negcolor{81.6} & \negcolor{82.7} & \negcolor{85.1} & EL & \negcolor{32.4} & \negcolor{33.0} & \negcolor{43.5} & \negcolor{84.0} \\
Wine & 32  & \negcolor{63.8} & \negcolor{64.6} & \negcolor{68.7} & NA & \negcolor{61.6} & \negcolor{64.6} & \negcolor{52.0} & \negcolor{64.8} \\
Wine & 512 & \negcolor{65.6} & \negcolor{69.7} & \negcolor{75.3} & EL & \negcolor{56.2} & \negcolor{56.4} & \negcolor{56.2} & \negcolor{71.0} \\
Yeast & 32  & \negcolor{43.1} & \negcolor{41.8} & \negcolor{41.8} & NA & \negcolor{36.7} & \negcolor{40.4} & \negcolor{33.7} & \negcolor{41.8} \\
Yeast & 512 & \negcolor{55.6} & \negcolor{58.2} & \negcolor{59.9} & EL & \negcolor{42.8} & \negcolor{36.7} & \negcolor{32.3} & \negcolor{59.3} \\
\bottomrule
\end{tabular}
\end{threeparttable}
\end{table}

\renewcommand{\negcolor}[1]{%
  \begingroup
  \dimen0=#1pt\relax
  \ifdim\dimen0<1.5pt \dimen0=100\dimen0\fi
  \ifdim\dimen0<50pt
    \cellcolor{ml-lime!10!white}{#1}
  \else\ifdim\dimen0<60pt
    \cellcolor{ml-lime!15!white}{#1}
  \else\ifdim\dimen0<70pt
    \cellcolor{ml-lime!20!white}{#1}
  \else\ifdim\dimen0<72pt
    \cellcolor{ml-lime!25!white}{#1}
  \else\ifdim\dimen0<73.5pt
    \cellcolor{ml-lime!35!white}{#1}
  \else\ifdim\dimen0<74.5pt
    \cellcolor{ml-lime!45!white}{#1}
  \else\ifdim\dimen0<75.5pt
    \cellcolor{ml-lime!55!white}{#1}
  \else\ifdim\dimen0<76.5pt
    \cellcolor{ml-lime!65!white}{#1}
  \else\ifdim\dimen0<80pt
    \cellcolor{ml-lime!75!white}{#1}
  \else\ifdim\dimen0<86pt
    \cellcolor{ml-lime!85!white}{#1}
  \else
    \cellcolor{ml-lime!85!white}{#1}
  \fi\fi\fi\fi\fi\fi\fi\fi\fi\fi
  \endgroup
}

\begin{table}[t]
\centering
\small
\setlength{\tabcolsep}{6pt}
\providecommand{\na}{\textcolor{black!50}{N/A}}
\caption{MMLU results for \textsc{MachineLearningLM} and baselines. (a) Macro accuracy across $M$-shot; (b) per-subject accuracies at 50-shot.}
\label{tab:mmlu_all}

\begin{subtable}{\linewidth}
\centering
\begin{threeparttable}
\begin{tabular}{lcccc}
\toprule
\textbf{Number of shots} & \textbf{Qwen-2.5-7B-Instruct} & \textbf{Qwen-2.5-7B} & \textbf{TabuLa-8B} & \textbf{\textsc{MachineLearningLM}} \\
\midrule
0   & \negcolor{73.8} & \negcolor{73.5} & \negcolor{61.6} & \negcolor{73.2} \\
10  & \negcolor{75.9} & \negcolor{75.5} & \negcolor{65.2} & \negcolor{75.1} \\
50  & \negcolor{75.8} & \negcolor{75.4} & \na              & \negcolor{75.4} \\
\bottomrule
\end{tabular}
\caption{Macro accuracy on the full MMLU benchmark across different $M$-shot settings at $temperature\!=\!0$. TabuLa-8B~\citep{gardner2024largescale} uses a maximum 8k-token length.}
\label{tab:mmlu_subject}
\end{threeparttable}
\end{subtable}

\vspace{0.6em} 

\begin{subtable}{\linewidth}
\centering
\begin{threeparttable}
\caption{Per-subject accuracies at 50-shot.}
\begin{tabular}{lcccc}
\toprule
\textbf{Subject} & \textbf{Qwen-2.5-7B-Inst.} & \textbf{Qwen-2.5-7B} & \textbf{TabuLa-8B} & \textbf{\textsc{{MachineLearningLM}}} \\
\midrule
high\_school\_statistics   & \negcolor{73.6} & \negcolor{69.9} & \negcolor{59.3} & \negcolor{74.1} \\
high\_school\_physics      & \negcolor{62.9} & \negcolor{55.6} & \negcolor{47.7} & \negcolor{61.6} \\
astronomy                  & \negcolor{86.8} & \negcolor{86.2} & \negcolor{68.4} & \negcolor{85.5} \\
college\_math       & \negcolor{47.0} & \negcolor{54.0} & \negcolor{32.0} & \negcolor{49.0} \\
college\_physics           & \negcolor{53.9} & \negcolor{61.8} & \negcolor{53.9} & \negcolor{57.8} \\
conceptual\_physics        & \negcolor{74.0} & \negcolor{74.0} & \negcolor{62.1} & \negcolor{76.2} \\
econometrics               & \negcolor{69.3} & \negcolor{64.9} & \negcolor{49.1} & \negcolor{65.8} \\
elementary\_math    & \negcolor{72.8} & \negcolor{74.1} & \negcolor{42.9} & \negcolor{71.4} \\
high\_school\_math  & \negcolor{55.6} & \negcolor{56.3} & \negcolor{32.6} & \negcolor{55.2} \\
\bottomrule
\end{tabular}
\begin{tablenotes}[para,flushleft]
\footnotesize
All accuracies (\%) are reported using \(50\) shots, except for TabuLa-8B, which suggests a maximum sequence length of 8k and therefore uses a 20-shot setting. Models are evaluated with \(temperature\!=\!0.05\), votes\(=\)3.
\end{tablenotes}
\end{threeparttable}
\label{tab:mmlu_subject_50shot}
\end{subtable}

\end{table}

\subsubsection{\textsc{MachineLearningLM} vs.\ tabular ICL models (TabICL / TabPFN)}
As Table~\ref{tab:shots_curve} shows, compared to state-of-the-art tabular ICL models like TabICL, \textsc{MachineLearningLM} can be modestly behind at very high shot counts—largely due to LLM compute requirements. Unlike tabular architectures with row/column attention and specialized tokenizer~\citep{hollmanntabpfn,qu2025tabicl,su2024tablegpt2,wang2021tuta}, \textsc{MachineLearningLM} uses a \emph{general-purpose} LLM backbone yet remains \emph{order-robust}: permuting in-prompt demonstrations leaves performance unchanged on average. Importantly, our method is \emph{LLM-compatible} by design and thus uniquely positioned to exploit \emph{external knowledge}, \emph{heterogeneous/multimodal inputs}, and \emph{reasoning-style} alignment (e.g., CoT), offering a practical path to close the remaining gap while retaining broad capabilities.

\subsubsection{\textsc{MachineLearningLM} in general chat workflows}
As shown in Table~\ref{tab:mmlu_all}, general abilities are \emph{well preserved}. On MMLU, our model achieves \(\sim\)73.2\% (0-shot) and \(\sim\)75.4\% (50-shot), comparable to strong general-purpose LLMs. Notably, we observe consistent gains in numeracy-heavy subjects (e.g., high-school statistics/conceptual physics). Although training LLMs on real tables, TabuLa-8B underperforms on MMLU (by \(\sim\)10\%), underscoring that our training strategy of reusing the model architecture and tokenizer integrates smoothly with general LLM capabilities and remains compatible with future multimodal/heterogeneous extensions.

\subsection{Empirical study}

\paragraph{Numeric-dominant tables.}
On pure numeric tables (e.g., \texttt{churn}, \texttt{maternal\_health\_risk}, \texttt{waveform}, \texttt{twonorm}, \texttt{vehicle}, \texttt{yeast}, \texttt{heloc}), our model remains competitive, indicating that the same prompt design handles both modalities without architecture changes.
 
\paragraph{Heterogeneous (text+numeric) tables.}
Our tabular encoding natively mixes free-text/categorical fields with numbers—\emph{without} mandatory text bucketing or learned text embeddings—yielding reliable gains on heterogeneous schemas. On mixed-feature datasets such as \texttt{bank}, \texttt{adult}, \texttt{credit-g}, \texttt{online\_shoppers}, \texttt{Bank\_Customer\_Churn\_Dataset}, and \texttt{okcupid\_stem}, \textsc{MachineLearningLM} consistently outperforms vanilla LLMs. \emph{By contrast}, for highly symbolic/abstract textual fields that behave more like token sequences than NL (e.g., DNA base strings in the UCI \texttt{splice} dataset), \textsc{MachineLearningLM} falls short relative to numerical modeling methods such as Random Forests.

\paragraph{High-cardinality labels.}
On tasks with many classes (e.g., \texttt{kropt}, \texttt{letter}~(26-way), \texttt{walking\_activity}), \textsc{MachineLearningLM} underperforms strong tabular baselines (RF/TabICL). We attribute this to pretraining that sampled tasks with $K \le 10$ classes, biasing the decoder toward small label vocabularies; when $K>10$, labels may become multi-token (e.g., ``12''). Expanding $K$ in synthesis is a promising mitigation. 

\paragraph{Resilience to class imbalance.}
Across the task suite, \textsc{MachineLearningLM} remains stable under skewed label distributions: its many-shot accuracy tracks Random-Forest–level performance while avoiding collapse-to-majority.  On notably imbalanced datasets—e.g., \texttt{bank}, \texttt{pc1}, and \texttt{kc1}—it sustains competitive accuracy.

\paragraph{Limits on forecasting-style tasks.}
On time-series/forecasting tasks---particularly the FOREX variants (\texttt{FOREX task series})---we observe significant declines relative to tabular methods. This gap is expected: our pretraining targets i.i.d.\ tabular prediction, whereas forecasting requires temporal inductive bias (lagged context, trend/seasonality priors) and higher numeric precision. We view this as an opportunity: future work will incorporate a time-aware strategy to better align \textsc{MachineLearningLM} with forecasting workloads. More limitations are detailed in Section ~\ref{sec:limitations}.

\section{Related Work}

\paragraph{Many-shot ICL and scaling laws.}
A growing literature observes that simply adding more demonstrations often yields diminishing or even negative returns in LLMs---e.g., one or a few high-quality demonstrations capture most gains~\citep{chen2023howmany,min-etal-2022-rethinking,liu2022makes,pan2023context}, long-context usage is position-sensitive such as ``lost in the middle''~\citep{liu2024lostinthemiddle,an2025why-ecl-falls-short,xiong-etal-2024-effective}, and many-shot gains can plateau in multimodal VLMs such as Flamingo beyond \(\sim\)32 shots~\citep{alayrac2022flamingo,tai2024linkcontext}. Recent efforts improve many-shot ICL via specialized training or mechanisms: DeepMind’s study shows task-dependent trends and proposes reinforced/unsupervised ICL variants~\citep{agarwal2024manyshot} to mitigate many-shot scaling; DrICL reweights demonstrations and modifies objectives to mitigate plateauing~\citep{zhang2025dricl}; semi-supervised many-shot ICL mitigates many-shot scaling when using self-generated annotations~\citep{gu2025scalinglaws}; multimodal works report mixed but often dataset-specific monotonic gains with longer contexts~\citep{jiang2024manyshotmm}. Moreover, long-context evaluation disentangles retrieval-like tasks from global context understanding, with many models degrading on the latter at \(16\)k tokens~\citep{zou-etal-2025-many}. 
TabDPT reports scaling laws with respect to both model size and pretraining-corpus size~\cite{ma2024tabdpt}. Unlike these approaches, \textsc{MachineLearningLM} induces many-shot ICL \emph{through pretraining on millions of synthetic tabular tasks}, yielding \emph{monotonic example-count scaling} while preserving the general reasoning of the base LLM.

\paragraph{LM-based ML learner.}
Compared with TabLLM \citep{hegselmann2023tabllm}—an approach that demands computationally intensive, task-specific fine-tuning and hyper-parameter tuning—\textsc{MachineLearningLM} adapts to new tabular tasks \emph{purely} through ICL at inference time, with no gradient updates. Crucially, it delivers markedly stronger reasoning over large volumes of numerical data, a long-standing weakness of TabLLM \footnote{\url{https://gael-varoquaux.info/science/carte-toward-table-foundation-models.html}}.
Most relevant to us, \citep{gardner2024largescale} proposes TabuLa-8B---finetuning Llama~3-8B on a web-scale corpus (T4) distilled from TabLib, and introducing a row-causal tabular masking (RCTM) scheme that packs samples by table and encourages few-shot behavior~\citep{gardner2024largescale}. Their main evaluations focus on \(k\!\in\![0,32]\) shots. 
Our synthetic pretraining complements real-world data pretraining and differs in three ways: (i) we \emph{continue pretraining} a general-purpose LLM \emph{without architectural modifications} without losing the backbone's general capabilities, (ii) we pretrain on \emph{synthetic, SCM-driven} tabular tasks with explicitly controlled diversity, ensuring no dataset leakage from downstream evaluations, and (iii) we target \emph{many-shot ICL} behavior on tabular tasks (1,024 demonstrations under a fixed context budget) using our proposed token-efficient method.

\paragraph{In-context tabular ML learners.}
TabPFN frames tabular classification as ICL with a pre-trained transformer hypernetwork and achieves strong few-shot accuracy on small tables~\citep{hollmanntabpfn}. TabICL scales PFN-style ICL to much larger tables via a two-stage, column-then-row architecture and synthetic pretraining~\citep{qu2025tabicl}. However, these tabular ICL models are trained independently of language models—incompatible with LM architectures or checkpoints—so they lack general text reasoning and open-domain knowledge, which limits application when textual fields carry meaning and constrains performance on multimodal (text–numeric) tasks.
In contrast, \textsc{MachineLearningLM} is built upon a pre-trained LLM backbone and therefore combines robust numeric processing with the ability to interpret textual headers, free-text cells, and world knowledge priors, offering a unified foundation model for heterogeneous, real-world in-context ML tasks.

\paragraph{Tool-using agents for ML vs.\ in-context learners.}
A complementary line of work evaluates or builds LLM agents that \emph{call external ML toolchains} (e.g., LightGBM/XGBoost/CatBoost, AutoML) to solve end-to-end ML engineering tasks, as in MLE-Bench~\citep{chan2024mlebench}, MLAgentBench~\citep{huang2023mlagentbench}, R\&D-Agent~\citep{yang2025rdagent}, and ML-Master~\citep{mlmaster2025}. However, their performance is bound by the invoked learners and pipelines. In contrast, \textsc{MachineLearningLM} explores a way to \emph{internalize} the learning procedure: the model performs in-context prediction, requiring neither per-task fine-tuning nor calls to external ML models, and \textsc{MachineLearningLM} can also be called by the MLE agent as a tool.

\section{Limitations}
\label{sec:limitations}
While \textsc{MachineLearningLM} shows strong in-context ML performance, several limitations remain.

\begin{itemize}
\item \textbf{Task scope.} Our pretraining corpus focuses on \emph{tabular classification} synthesized from SCMs. We do not yet cover regression, ranking, time–series forecasting, or structured prediction, and we cap the number of classes ($K\!\le\!10$). Extending beyond IID rows (e.g., temporal or relational dependencies) is future work.

\item \textbf{Context length and compute.} Continued pretraining used a 32k-token context; although inference generalizes beyond this (tested up to $\sim$131k tokens), scaling to many thousands of shots like TabPFN/TabICL remains challenging due to compute and memory costs.

\item \textbf{Numerical encoding trade-offs.} The $[0,999]$ integer mapping preserves order but \emph{coarsens} magnitudes and can obscure semantically meaningful constants (e.g., age 18). As context windows grow, more expressive number encodings can be explored to retain such cues while remaining token-efficient.

\item \textbf{Warm-start bias.} The RF-mimic warm-up and example-level consensus filtering may bias early training toward tree-like decision boundaries and easy examples. Although we disable consensus after warm-up, some inductive bias may persist; we currently distill \emph{labels only} (not rationales). How to best leverage external teachers is still an open question.

\item \textbf{Model scale and adaptation.} Results are with a 7B backbone and low-rank adaptation (LoRA rank 8). Larger backbones, alternative adapters, or optimizer/regularization choices may further improve many-shot numeracy but were not explored here.
\end{itemize}

\section{Future Work}
\label{future}
We hope \textsc{MachineLearningLM} will ignite new lines of research, as exemplified below.
Feel free to contact us for further discussions\,{\textcolor{msGreen}\faHandshake}.

\begin{itemize}

\item \textbf{Beyond text and numbers: synthetic tasks with multimodal features.}
Starting from our synthetic tabular tasks, map numerical/categorical features to \emph{weakly labeled} multimodal surrogates—short texts, speech snippets, images,  geospatial signals, or time series—using generative back-ends (LLM/VLM/TTS) conditioned on latent factors to preserve coarse class semantics. This produces controllable, cross-modal pairs aligned with the tabular schema~\footnote{Table formats (e.g., HTML tables) naturally organize images, hyperlinks, and embedded objects.} and enables \textsc{MachineLearningLM} to practice multimodal in-context prediction on heterogeneous signals.

\item \textbf{Longer contexts via parallelism and system optimizations.}
Extend context length using tensor/pipeline parallelism and memory-efficient attention/KV caching to support substantially more in-prompt examples. Recent work on cache-augmented architectures demonstrates that principled KV-cache management can significantly improve efficiency and scalability for long-context reasoning~\citep{bhaskar2025cache}, suggesting promising directions to integrate such mechanisms into \textsc{MachineLearningLM}.

\item \textbf{Toward interpretability: narrative distillation and reasoning-augmented learning.}\\
Each task in our pretraining corpus is generated from a structural causal model (SCM), which we treat as an intrinsic source of explanation. We train the LM to \emph{narrate} the underlying SCM—variables, relations, and mechanisms—as an auxiliary objective aligned with the prediction task, thereby yielding more faithful rationales and stronger reasoning.
Moreover, couple feature attribution signals (e.g., SHAP~\citep{lundberg2017shap} / LIME~\citep{ribeiro2016lime}) with \textsc{MachineLearningLM}’s generation to produce faithful instance- and cohort-level narratives, while distilling rules from ensembles (e.g., inTrees~\citep{deng2014intrees} / RuleFit~\citep{friedman2008rulefit}) into compact, human-readable reasoning traces. Recent advances such as DeepSeek-R1~\citep{deepseek-r1} demonstrate the promise of incentivizing reasoning capability in LLMs through RL. This integration promises not only accurate predictions but also transparent rationales, enhancing compliance and trust.

\item \textbf{Uncertainty-aware responses.}
Recent analyses from OpenAI argue that training/evaluation schemes which \emph{penalize} uncertainty and implicitly \emph{reward} guessing are a root cause of hallucinations~\footnote{\url{https://openai.com/index/why-language-models-hallucinate/}}. We suspect a similar incentive mismatch in our setting. Future work may (i) train on \emph{teacher predictive distributions}—e.g., randomized/ensemble teachers that provide calibrated confidences—instead of hard labels, optimized with proper scoring rules (NLL/Brier) and selective-prediction losses that allow abstention; (ii) explicitly represent and reward well-calibrated uncertainty in the output schema (e.g., an \texttt{UNCERTAIN}/defer option with risk–coverage evaluation); and (iii) directly optimize task-level metrics (AUROC, RMSE, F1) via reinforcement learning or differentiable surrogates, with robust reward shaping (e.g., clipped/Huberized or quantile-based rewards) to reduce outlier sensitivity and improve end-to-end utility.

\item \textbf{Combined with Retrieval-Augmented Methods.}  
Retrieval-augmented method has shown that attaching a retrieval module to LLMs enables scalable any-shot learning and yields power-law error reductions as data grows \citep{wen2025ragtabicl}. Integrating such retrieval into \textsc{MachineLearningLM} can boost many-shot example limits by dynamically injecting the most relevant examples during pretraining and inference.

\item \textbf{Real-data alignment via continued fine-tuning.} Beyond purely synthetic pretraining, we will perform lightweight continued fine-tuning on a curated suite of real tabular prediction tasks to better align with practical distributions.

\item \textbf{Extending \textsc{MachineLearningLM} to fully leverage Agent Memory.}
Recent advances leverage memory mechanisms so that agents can recall and reuse successful or similar past experiences to inform their next action~\citep{wang2023voyager,lin2025se,wang2024agent}. Such experiences may be textual traces, numerical features, or temporal signals. By pretraining the policy model with many-shot in-context experiences, the agent can improve decision-making and robustness in dynamic environments.

\item \textbf{Integration with MLE agents.}
Fuse \textsc{MachineLearningLM} with MLE-agent–style planners to orchestrate end-to-end MLE workflows. The agent treats \textsc{MachineLearningLM} both as a solver and as an internal tool, enabling a flexible interface with heterogeneous real-world MLE tasks.

\end{itemize}

\section{Conclusion}

We introduced \textsc{MachineLearningLM}, a portable continued-pretraining recipe that equips a general-purpose LLM with robust in-context tabular prediction \emph{without} architectural or tokenizer changes. By synthesizing millions of SCM-driven tasks and using a Random-Forest warm start, the model acquires strong numerical modeling while preserving general knowledge and reasoning. A token-efficient tabular encoding, integer-based number normalization, and sequence-level batch-prediction together expand the effective context budget and amortize prompt overhead. Across diverse out-of-distribution classification tasks, \textsc{MachineLearningLM} remarkably improves over its backbone and approaches random-forest–level accuracy from 8 to 512 shots; accuracy further increases up to 1{,}024 shots and is robust to exemplar-order permutations. These results indicate that targeted continued pretraining on synthetic prediction tasks is a practical path to scaling many-shot ICL within general-purpose LLMs. Finally, we discuss the current limitations and chart several promising avenues for future work.

\section*{Author Contributions}
\label{sec:author}
\textbf{Conceptualization} (\emph{LLM continued pretraining with SCM-based synthesis, RF-mimic warm-start, token-efficient prompting, and order-robust self-consistency}): Haoyu Dong\footnote{University of Chinese Academy of Sciences, Microsoft, \url{donghaoyu82@gmail.com}}.

\textbf{Methodology}: Pretraining recipe and objectives --- Haoyu Dong, Pengkun Zhang\footnote{South China University of Technology, \url{sezhangpengkun@mail.scut.edu.cn}}; SCM task synthesis --- Pengkun Zhang, Haoyu Dong; RF-mimic warm-start (gating and consensus filtering) --- Haoyu Dong; token-efficient prompting --- Haoyu Dong; self-consistency --- Yanzhen Shen\footnote{Stanford University, \url{yanzhen4@stanford.edu}}, Mingzhe Lu\footnote{University of Chinese Academy of Sciences, \url{lumingzhe23@mails.ucas.edu.cn}}, Haoyu Dong.

\textbf{Software \& Engineering}: Pretraining pipeline and training scripts --- Pengkun Zhang, Haoyu Dong; evaluation harness --- Yanzhen Shen, Mingzhe Lu, Haoyu Dong; SCM generator and data pipelines --- Pengkun Zhang, Haoyu Dong; token-efficient prompting utilities --- Haoyu Dong; code optimization, cross-platform support, and parallelization --- Mingzhe Lu, Pengkun Zhang.

\textbf{Investigation \& Validation}: Experiments and statistical aggregation of results --- Mingzhe Lu, Pengkun Zhang, Haoyu Dong.

\textbf{Writing}: Original draft (\emph{main text, tables, and figures}) --- Haoyu Dong; review and editing --- Yanzhen Shen, Pengkun Zhang, Mingzhe Lu, Guolin Ke\footnote{Individual, \url{guolin.ke@outlook.com}}.

\textbf{Limitations \& Future Work}: Haoyu Dong, Guolin Ke.

\textbf{ML Expertise}: High-level guidance on ML practice, interpretability, and online/continual learning directions --- Guolin Ke.

Moreover, we thank Hui Xue for suggestions on method and evaluation, including adding the MMLU benchmark and future directions that leverage random-forest predictive probabilities.

\bibliography{iclr2026_conference}
\bibliographystyle{iclr2026_conference}

\clearpage
\appendix

\section{List of Symbols}
\label{app:symbols}
\begingroup\small
\begin{longtable}{@{}Q{0.16\linewidth} P{0.6\linewidth} P{0.16\linewidth}@{}}
\toprule
\textbf{Symbol} & \textbf{Meaning} & \textbf{Range/Default} \\
\midrule
\endfirsthead
\toprule
\textbf{Symbol} & \textbf{Meaning} & \textbf{Range/Default} \\
\midrule
\endhead

$\mathcal{G}$ & DAG of the structural causal model (SCM) & — \\
$v$ & Node index in $\mathcal{G}$ & — \\
$x_v$ & Variable value at node $v$ & — \\
$f_v$ & Structural function at node $v$ & — \\
$\varepsilon_v$ & i.i.d.\ noise term & — \\

$M$ & \# in-context demonstrations (shots) & $\le 1024$ \\
$N$ & \# query/test rows predicted per batched sequence & $50$  \\
$d$ & Feature dimension (columns) & $5\sim 50$ \\
$K$ & \# classes & $2\sim 10$ \\

$S$ & Demonstration set $\{s_i\}_{i=1}^{M}$ & — \\
$Q$ & Query set $\{x^{(i)}\}_{i=1}^{N}$ & — \\
$y$ & Target JSON array of predictions & — \\
$\hat{\ell}_i$ & Predicted label for query $i$ & — \\

$\mathrm{tok}(\cdot)$ & Tokenizer & — \\
$T$ & Target token length & — \\
$\theta$ & Model parameters & — \\
$\mathcal{L}(\theta)$ & Training loss (NLL) & — \\

$H$ & Token cost of the instruction/schema header & — \\
$R$ & Per-row token cost (demo or query) & — \\
$C$ & Amortized token cost per predicted label & — \\
$z$ & $z$-score normalized numeric value & — \\
$i$ & Integer after $[0,999]$-norm mapping & — \\

$p_0$ & Conservative random baseline accuracy &  — \\
$N_{\text{correct}}$ & \# correct predictions by RF per batched sequence & — \\
$\alpha$ & One-sided binomial test threshold & $0.2$ \\
$\kappa$ & Cohen’s $\kappa$ (chance-corrected agreement) & $>0.01$  \\
$\delta_{\text{bacc}}$ & Balanced-accuracy margin & $0.03$  \\
$\delta_{\text{F1}}$ & Macro-F1 margin & $0.00$  \\
$\tau_{\text{dom}}$ & Max dominant-class fraction & $0.95$  \\

$V$ & \# shuffled prompt variants & $5$ \\
$p_i(y_j)$ & Next-token prob.\ for label $y_j$ under variant $P_i$ & — \\
$\tilde p(y_j)$ & Aggregated prob.: $\sum_{i=0}^{V-1} p_i(y_j)$ & — \\

\bottomrule
\end{longtable}
\endgroup
\clearpage

\section{Many-shot Scaling Per-task evaluation}
\label{extend_shot_eval}

\small
\setlength{\tabcolsep}{6pt}
\begin{longtable}{l r c c c c}
\caption{Per-dataset test accuracy from 8 to 1,024 shots on more sampled domain-representative datasets in Talent. Abbreviations: Qwen-2.5-7B $=$ Qwen\mbox{-}2.5\mbox{-}7B\mbox{-}Instruct; Our $=$ \textsc{MachineLearningLM} (base: Qwen\mbox{-}2.5\mbox{-}7B\mbox{-}Instruct).}\\
\toprule
\textbf{Dataset} & \textbf{Shots} & \textbf{KNN} & \textbf{RF} & \textbf{Qwen-2.5-7B} & \textbf{Our} \\
\midrule
\endfirsthead
\multicolumn{6}{c}{\tablename\ \thetable{} -- continued from previous page} \\
\toprule
\textbf{Dataset} & \textbf{Shots} & \textbf{KNN} & \textbf{RF} & \textbf{Qwen-2.5-7B} & \textbf{Our} \\
\midrule
\endhead
\midrule \multicolumn{6}{r}{Continued on next page} \\
\endfoot
bank & 8 & 0.8657 & 0.8695 & 0.8526 & 0.8704 \\
bank & 16 & 0.8723 & 0.8759 & 0.8666 & 0.8707 \\
bank & 32 & 0.8725 & 0.8764 & 0.8722 & 0.8727 \\
bank & 64 & 0.8764 & 0.8828 & 0.8780 & 0.8771 \\
bank & 128 & 0.8778 & 0.8847 & 0.8798 & 0.8830 \\
bank & 256 & 0.8757 & 0.8873 & 0.8803 & 0.8867 \\
bank & 512 & 0.8762 & 0.8897 & 0.8808 & 0.8870 \\
bank & 1,024 & 0.8801 & 0.8947 & 0.8795 & 0.8903 \\
\bottomrule
\end{longtable}

\section{Extended Per-task Evaluation}
\label{extend_eval}

\small
\setlength{\tabcolsep}{6pt}
\begin{longtable}{l r c c c c}
\caption{Per-dataset test accuracy from 8 to 512 shots on extended datasets in Talent. Evaluation on all 8-1024 shots and full datasets is very computation intensive, and we leave it for the next revision. Abbreviations: Qwen $=$ Qwen\mbox{-}2.5\mbox{-}7B\mbox{-}Instruct; Our $=$ \textsc{MachineLearningLM} (base: Qwen\mbox{-}2.5\mbox{-}7B\mbox{-}Instruct).}\\
\toprule
\textbf{Dataset} & \textbf{Shots} & \textbf{KNN} & \textbf{RF} & \textbf{Qwen} & \textbf{Our} \\
\midrule
\endfirsthead
\multicolumn{6}{c}{\tablename\ \thetable{} -- continued from previous page} \\
\toprule
\textbf{Dataset} & \textbf{Shots} & \textbf{KNN} & \textbf{RF} & \textbf{Qwen} & \textbf{Our} \\
\midrule
\endhead
\midrule \multicolumn{6}{r}{Continued on next page} \\
\endfoot
\bottomrule
\endlastfoot
Amazon\_employee\_access & 8 & 0.9254 & 0.9245 & 0.8264 & 0.9227 \\
Amazon\_employee\_access & 16 & 0.9221 & 0.9274 & 0.813 & 0.9257 \\
Amazon\_employee\_access & 32 & 0.928 & 0.9311 & 0.8407 & 0.9279 \\
Amazon\_employee\_access & 64 & 0.928 & 0.9295 & 0.8895 & 0.9266 \\
Amazon\_employee\_access & 128 & 0.9305 & 0.9293 & 0.91 & 0.9343 \\
Amazon\_employee\_access & 256 & 0.9276 & 0.9313 & 0.9215 & 0.9364 \\
Amazon\_employee\_access & 512 & 0.9291 & 0.9305 & 0.9291 & 0.939 \\
BNG\_breast\_w & 8 & 0.7732 & 0.8881 & 0.7915 & 0.921 \\
\end{longtable}

\end{document}